\documentclass{article}

\usepackage{microtype}

\usepackage{hyperref}
\usepackage{url}

\usepackage[accepted]{icml2020}

\usepackage{enumerate, amsmath, amsthm, amssymb, graphicx, pifont, csquotes, dashrule, mathrsfs, tikz, bbm, booktabs, bm, verbatim}
\usepackage[framemethod=TikZ]{mdframed}

\usepackage{caption}
\usepackage{subcaption}
\usepackage{wrapfig}
\usepackage{adjustbox}
\usepackage{sidecap}

\newcommand{\pie}{\pi^*}

\newif\ifsubmit
\submitfalse
\ifsubmit
\newcommand{\dnote}[1]{}
\newcommand{\deynote}[1]{}
\newcommand{\alekh}[1]{}
\else
\newcommand{\dnote}[1]{\textcolor{blue}{Dilip: #1}}
\newcommand{\deynote}[1]{\textcolor{orange}{Dey: #1}}
\newcommand{\alekh}[1]{{\color{magenta} Alekh: #1}}
\fi

\icmltitlerunning{Reparameterized Variational Divergence Minimization for Stable Imitation}

\begin{document}

\twocolumn[
\icmltitle{Reparameterized Variational Divergence Minimization for Stable Imitation}



\icmlsetsymbol{equal}{*}

\begin{icmlauthorlist}
\icmlauthor{Dilip Arumugam}{st}
\icmlauthor{Debadeepta Dey}{msr}
\icmlauthor{Alekh Agarwal}{msr}
\icmlauthor{Asli Celikyilmaz}{msr}
\icmlauthor{Elnaz Nouri}{msr}
\icmlauthor{Bill Dolan}{msr}
\end{icmlauthorlist}

\icmlaffiliation{st}{Department of Computer Science, Stanford University, Stanford, California, USA}
\icmlaffiliation{msr}{Microsoft Research, Redmond, Washington, USA}

\icmlcorrespondingauthor{Dilip Arumugam}{dilip@cs.stanford.edu}
\icmlcorrespondingauthor{Debadeepta Dey}{dedey@microsoft.com}

\icmlkeywords{Machine Learning, ICML}

\vskip 0.3in
]



\printAffiliationsAndNotice{}  

\begin{abstract}
While recent state-of-the-art results for adversarial imitation-learning algorithms are encouraging, recent works exploring the imitation learning from observation (ILO) setting, where trajectories \textit{only} contain expert observations, have not been met with the same success. Inspired by recent investigations of $f$-divergence manipulation for the standard imitation learning setting~\citep{ke2019imitation,kamyar2019adiv}, we here examine the extent to which variations in the choice of probabilistic divergence may yield more performant ILO algorithms. We unfortunately find that $f$-divergence minimization through reinforcement learning is susceptible to numerical instabilities. We contribute a reparameterization trick for adversarial imitation learning to alleviate the optimization challenges of the promising $f$-divergence minimization framework. Empirically, we demonstrate that our design choices allow for ILO algorithms that outperform baseline approaches and more closely match expert performance in low-dimensional continuous-control tasks. 
\end{abstract}

\section{Introduction}

Imitation Learning (IL)~\citep{osa2018algorithmic} is a paradigm of reinforcement learning~\citep{sutton1998introduction} in which the agent has access to an optimal, reward-maximizing expert for the underlying environment. This access is usually provided via a dataset of trajectories where each observed state is annotated with the action prescribed by the expert policy. This is a powerful learning paradigm in contrast to standard reinforcement learning since not all tasks of interest admit easily-specified reward functions. Additionally, not all environments are amenable to the prolonged and potentially unsafe exploration needed for reward-maximizing agents to arrive at satisfactory policies~\citep{Achiam2017ConstrainedPO,Chow2019LyapunovbasedSP}. 


\begin{figure}
    \centering
    \includegraphics[width=\linewidth]{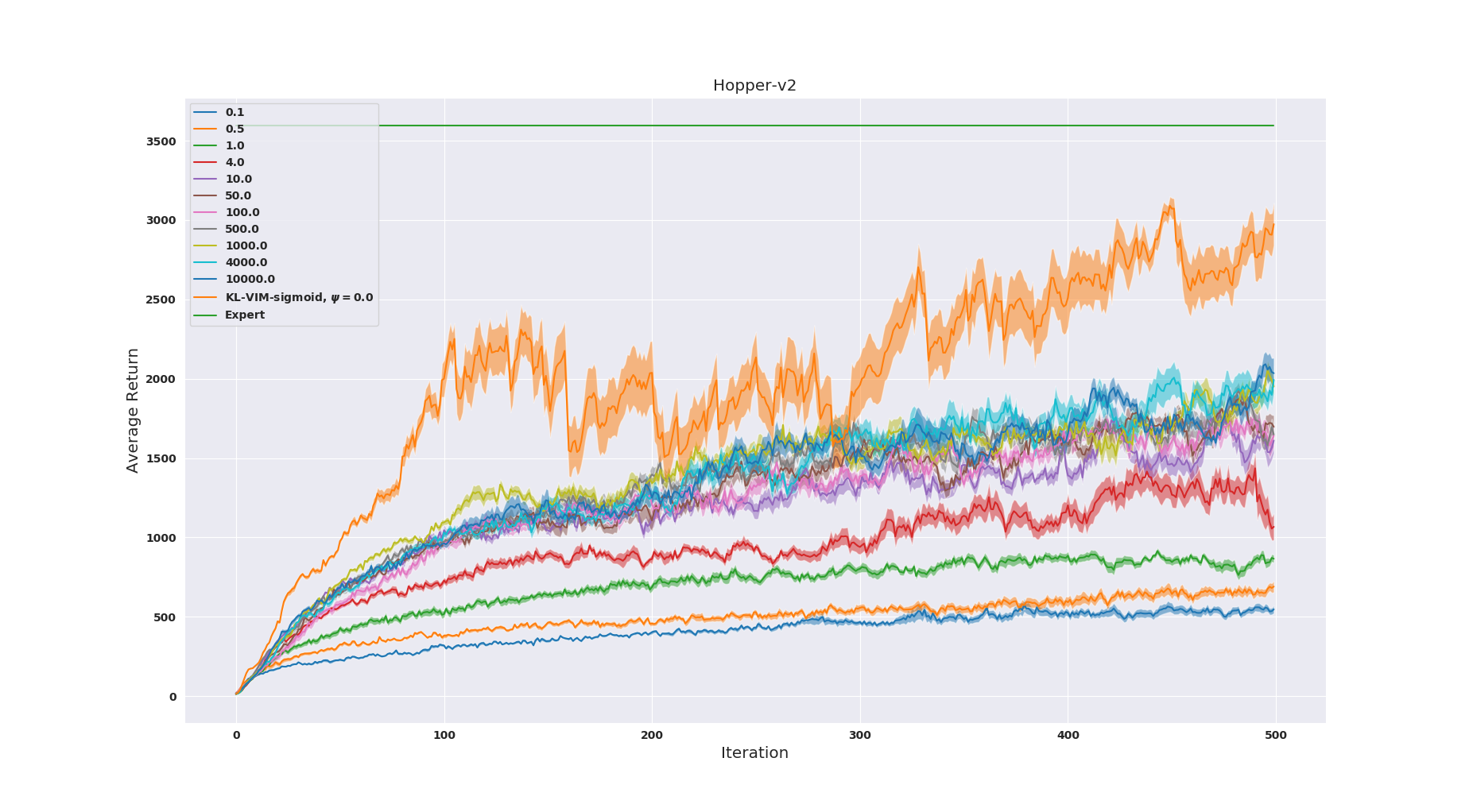}
    \caption{Comparing our reparameterized adversarial imitation-learning algorithm (orange) utilizing the KL divergence against that of \citet{ke2019imitation}, varying gradient norm clipping thresholds to account for numerical instability. We call attention to the parameter sensitivity of the latter approach that, despite tuning, still produces a weaker imitation policy.}
    \label{fig:kl_stability_hopper}
\end{figure}

While the traditional formulation of the IL problem assumes access to optimal expert action labels, the provision of such information can often be laborious (in the case of a real, human expert) or incur significant cost due to elaborate instrumentation needed to record expert actions. Additionally, this restrictive assumption removes a vast number of rich, observation-only data sources from consideration~\citep{zhou2018towards}. To bypass these challenges, recent work~\citep{Liu2018ImitationFO,torabi2018behavioral,torabi2018generative,edwards2019imitating,sun2019provably} has explored a more natural problem formulation in which an agent must recover an imitation policy from a dataset containing only expert observation sequences. While this Imitation Learning from Observations (ILO) setting carries tremendous potential, such as enabling an agent to learn complex tasks from watching freely available videos on the Internet, it also is fraught with significant additional challenges. In this paper, we show how to incorporate insights from the generative adversarial networks literature to advance the state-of-the-art in ILO.

The rich literature on Generative Adversarial Networks~\citep{goodfellow2014generative} has expanded in recent years to include alternative formulations of the underlying objective that yield qualitatively different solutions to the saddle-point optimization problem~\citep{li2015generative,dziugaite2015training,zhao2016energy,nowozin2016f,arjovsky2017wasserstein,Gulrajani2017ImprovedTO}. Of notable interest are the findings of \citet{nowozin2016f} who present Variational Divergence Minimization (VDM), a generalization of the generative-adversarial approach to arbitrary choices of distance measures between probability distributions drawn from the class of $f$-divergences~\citep{ali1966general,csiszar2004information}. Applying VDM with varying choices of $f$-divergence, \citet{nowozin2016f} encounter learned synthetic distributions that can exhibit differences from one another while producing equally realistic samples. Translating this idea for imitation is complicated by the fact that the optimization of the generator occurs via policy-gradient reinforcement learning~\citep{sutton2000policy}. Existing work in combining adversarial IL and $f$-divergences~\citep{ke2019imitation,kamyar2019adiv} either fails to account for this difference or engages in an extensive hyperparameter search over thresholding and gradient clipping parameters to elude numerical instability. In the former case, the end results are IL algorithms that scale poorly to environments with higher-dimensional observations, as partially shown in Figure \ref{fig:kl_stability_hopper}, where the effort of a tedious hyperparameter search still fails to deliver on a strong imitation policy. 

\textbf{The primary contribution of this work is a reparameterization scheme for stabilizing adversarial ILO methods.} As a consequence of this stabilization, we are able to investigate the VDM principle and alternative $f$-divergences in detail. We find that different choices of $f$-divergence, coupled with discriminator regularization, can improve upon the existing state-of-the-art in ILO on certain problems.
For ease of exposition, we begin by first examining the standard IL problem and outline a reparameterization of the $f$-VIM framework of \citet{ke2019imitation} (Sections \ref{sec:fvim} \& \ref{sec:fvim_refac}). Our version transparently exposes the choices practitioners must make when designing adversarial IL algorithms for arbitrary choices of $f$-divergence. While this enables many potential avenues for future exploration, all of our experiments focus on a single instantiation that allows for stable training of good imitation policies across multiple choices of $f$-divergence. We then return to the ILO setting and introduce $f$-VIMO for adversarial ILO algorithms under arbitrary $f$-divergences (Section \ref{sec:fvimo}). We conclude by examining the efficacy of varying $f$-divergences across a range of continuous-control tasks in the MuJoCo~\citep{todorov2012mujoco} domain (Section \ref{sec:exps}). 

Our empirical results validate our framework as a viable unification of adversarial ILO methods under the VDM principle. We also find that under the appropriate choice of $f$-divergence, recent advances in stabilizing regularization for adversarial training~\citep{mescheder2018training} can help to improve performance (Section \ref{sec:discr_reg}). Nevertheless, there is still a significant performance gap between the recovered imitation policies and expert behavior for tasks with sufficiently high-dimensional observations, motivating future empirical and theoretical work on ILO algorithms. 

\section{Related Work}

The algorithms presented in this work fall in with inverse reinforcement learning (IRL)~\citep{ng2000algorithms,Abbeel2004ApprenticeshipLV,Syed2007AGA,ziebart2008maximum,finn2016guided,ho2016generative,fu2018learning} approaches to IL. 
Our work focuses attention exclusively on adversarial methods for their widespread effectiveness across a range of imitation tasks without requiring interactive experts~\citep{ho2016generative,li2017infogail,fu2018learning,Kostrikov2018DiscriminatorActorCriticAS}; at the heart of these methods is the Generative Adversarial Imitation Learning (GAIL)~\citep{ho2016generative} approach which produces high-fidelity imitation policies and achieves state-of-the-art results across numerous continuous-control benchmarks by leveraging the expressive power of Generative Adversarial Networks (GANs)~\citep{goodfellow2014generative} for modeling complex distributions over a high-dimensional support. 
Recognizing that the probabilistic divergence used to compute the distance between these distributions is a free parameter~\citep{nowozin2016f}, the goal of this work is to demonstrate the effect of changing $f$-divergences on the performance of ILO algorithms.

While a large body of prior work exists for IL, recent work has drawn attention to the more challenging problem of imitation learning from observation~\citep{Sermanet2017TimeContrastiveNS,Liu2018ImitationFO,Goo2018OneShotLO,Kimura2018InternalMF,torabi2018behavioral,torabi2018generative,edwards2019imitating,sun2019provably}. 
Many early approaches to ILO use expert observation sequences to learn a semantic embedding space so that distances between observation sequences of the imitation and expert policies can serve as a cost signal to be minimized via reinforcement learning~\citep{Gupta2017LearningIF,Sermanet2017TimeContrastiveNS,Dwibedi2018LearningAR,Liu2018ImitationFO}. In contrast, \citet{torabi2018behavioral} introduce Behavioral Cloning from Observation (BCO) which leverages state-action trajectories collected under a random policy to train an inverse dynamics model for inferring the action responsible for a transition between two input states (assuming the two represent a state and next-state pair). With this inverse model in hand, the observation-only demonstration data can be converted into the more traditional dataset of state-action pairs over which standard BC can be applied. Recognizing the previously discussed limitations of BC approaches, \citet{torabi2018generative} introduce the natural GAIL counterpart for ILO, Generative Adversarial Imitation from Observation (GAIFO); GAIFO is identical to GAIL except the distributions under consideration in the adversarial game are over state transitions (state and next-state pairs), as opposed to state-action pairs requiring expert action labels. While \citet{torabi2018generative} offer empirical results for continuous-control tasks with low-dimensional features as well as raw image observations, GAIFO falls short of expert performance in both settings leaving an open challenge for scalable ILO algorithms that achieve expert performance across a wide spectrum of tasks. Following the insights laid out in \citet{ke2019imitation,kamyar2019adiv} for the standard IL setting, the algorithms we present generalize GAIFO to arbitrary $f$-divergences. 
For a more in-depth survey of ILO approaches, we refer readers to \citet{Torabi2019RecentAI}. We refer readers to the Appendix for a more complete overview of prior work.

\section{Background}

We begin by formulating the problems of IL and ILO respectively before connecting them to $f$-divergences and VDM.

\subsection{Imitation Learning}

We operate within the Markov Decision Process (MDP) formalism~\citep{bellman1957markovian,puterman2014markov} defined as a five-tuple $\mathcal{M} = \langle \mathcal{S}, \mathcal{A}, \mathcal{R}, \mathcal{T}, \gamma \rangle$ where $\mathcal{S}$ denotes a (potentially infinite) set of states, $\mathcal{A}$ denotes a (potentially infinite) set of actions, $\mathcal{R}:\mathcal{S} \times \mathcal{A} \times \mathcal{S} \rightarrow \mathbb{R}$ is a reward function, $\mathcal{T}:\mathcal{S} \times \mathcal{A} \rightarrow \Delta(\mathcal{S})$ is a transition function, and $\gamma \in [0,1)$ is a discount factor. At each timestep, the agent observes the current state of the world, $s_t \in \mathcal{S}$, and randomly samples an action according to its stochastic policy $\pi: \mathcal{S} \rightarrow \Delta(\mathcal{A})$. The environment then transitions to a new state according to the transition function $\mathcal{T}$ and produces a reward signal according to the reward function $\mathcal{R}$ that is communicative of the agent's progress through the overall task.

Unlike, the traditional reinforcement learning paradigm, the decision-making problem presented in IL lacks a concrete reward function; in lieu of $\mathcal{R}$, a learner is provided with a dataset of expert demonstrations $\mathcal{D} = \{\tau_1, \tau_2, \ldots \tau_N\}$ where each $\tau_i = (s_{i1}, a_{i1}, s_{i2}, a_{i2}, \ldots)$ represents the sequence of states and corresponding actions taken by an expert policy, $\pie$. Naturally, the goal of an IL algorithm is to synthesize a policy $\pi$ using $\mathcal{D}$, along with access to the MDP $\mathcal{M},$ whose behavior matches that of $\pie$.

While there are several possible avenues for using $\mathcal{D}$ to arrive at a satisfactory imitation policy, our work focuses on adversarial methods that build around GAIL~\citep{ho2016generative}. Following from the widespread success of GANs~\citep{goodfellow2014generative}, GAIL offers a highly-performant approach to IL wherein transitions iteratively sampled from the current imitation policy are first used to update a discriminator, $D_\omega(s,a)$, acting as a binary classifier between state-action pairs sampled according to the distributions induced by the expert and student. Subsequently, treating the imitation policy as a generator, policy-gradient reinforcement learning~\citep{sutton2000policy} is used to shift the current policy towards expert behavior, issuing higher rewards for those generated state-action pairs that are regarded as belonging to the expert according to $D_\omega(s,a)$.  More formally, this minimax optimization follows as
\begin{multline}
    \min\limits_\pi \max\limits_\omega \mathbb{E}_{(s,a) \sim \rho^{\pie}}[\log(D_\omega(s,a))] \\ + \mathbb{E}_{(s,a) \sim \rho^{\pi}}[\log(1 - D_\omega(s,a))]
    \label{eq:gail_obj}
\end{multline}

where $\rho^{\pie}(s,a)$ and $\rho^{\pi}(s,a)$ denote the undiscounted stationary distributions over state-action pairs for the expert and imitation policies respectively. Here $D_\omega(s,a) = \sigma(V_\omega(s,a))$ where $V_\omega(s,a)$ represents the unconstrained output of a discriminator neural network with parameters $\omega$ and $\sigma(v) = (1 + e^{-x})^{-1}$ denotes the sigmoid activation function. Since the imitation policy only exerts control over the latter term in the above objective, the per-timestep reward function maximized by reinforcement learning is given as $r(s,a,s') = -\log(1 - D_\omega(s,a))$. In practice, an entropy regularization term is often added to the objective when optimizing the imitation policy so as to avoid premature convergence to a suboptimal solution~\citep{Mnih2016AsynchronousMF,ho2016generative,neu2017unified}.

\subsection{Imitation Learning from Observation}

In order to accommodate various observation-only data sources~\citep{zhou2018towards} and remove the burden of requiring expert action labels, the ILO setting adjusts the expert demonstration dataset $\mathcal{D}$ such that each trajectory $\tau_i = (s_{i1}, s_{i2}, \ldots)$ consists only of expert observation sequences. Retaining the goal of recovering an imitation policy that closely resembles expert behavior, \citet{torabi2018generative} introduce GAIFO as the natural extension of GAIL for matching the state transition distribution of the expert policy. Note that an objective for matching the stationary distribution over expert state transitions enables the provision of per-timestep feedback while simultaneously avoid the issues of temporal alignment that arise when trying to match trajectories directly~\citep{finn2016connection,fu2018learning}. The resulting algorithm iteratively finds a solution to the following minimax optimization:

\begin{multline}
    \min\limits_\pi \max\limits_\omega \mathbb{E}_{(s,s') \sim \rho^{\pie}}[\log(D_\omega(s,s'))] \\ + \mathbb{E}_{(s,s') \sim \rho^{\pi}}[\log(1 - D_\omega(s,s'))]
    \label{eq:gaifo_obj}
\end{multline}

where $\rho^{\pie}(s,s')$ and $\rho^{\pi}(s,s')$ now denote the analogous stationary distributions over successive state pairs while $D_\omega(s,s') = \sigma(V_\omega(s,s'))$ represents binary classifier over state pairs. Similar to GAIL, the imitation policy is optimized via policy-gradient reinforcement learning with per-timestep rewards computed according to $r(s,a,s') = -\log(1 - D_\omega(s,s'))$ and using entropy regularization as needed.

\section{Approach}
\label{sec:approach}

In this section, we begin with an overview of $f$-divergences, their connection to GANs, and their impact on IL through the $f$-VIM framework~\citep{ke2019imitation} (Section \ref{sec:fvim}). We then present our reparameterization of the framework for the IL setting (Section \ref{sec:fvim_refac}) before extending to the ILO setting (Section \ref{sec:fvimo}). We conclude with discussion of the regularization technique used to stabilize discriminator training in our experiments (Section \ref{sec:discr_reg}).

\begin{table*}[h]
\centering
\begin{adjustbox}{width=\textwidth}
\small
\begin{tabular}{cccccc}
\toprule
    Name & Output Activation $g_f$ & $\text{dom}_{f^*}$ & Conjugate $f^*(t)$ & $\text{dom}_{f^{*-1}}$ & Conjugate Inverse $f^{*-1}(t)$\\ \hline
    Total Variation (TV) & $\frac{1}{2}\tanh(v)$ & $-\frac{1}{2} \leq t \leq \frac{1}{2}$ & $t$ & $-\frac{1}{2} \leq t \leq \frac{1}{2}$ & $t$\\
    Kullback-Leibler (KL) & $v$ & $\mathbb{R}$ & $\exp(t-1)$ & $\mathbb{R}_+$ & $1 + \log(t)$ \\
    Reverse KL (RKL) & $-\exp(v)$ & $\mathbb{R}_-$ & $-1-\log(-t)$ & $\mathbb{R}$ & $-\exp(-1-t)$\\
    GAN & $-\log(1 + \exp(-v))$ & $\mathbb{R}_-$ & $-\log(1 - \exp(t))$ & $\mathbb{R}_+$ & $\log(1 - \exp(-t))$ \\
\end{tabular}
\end{adjustbox}
\caption{Table of various $f$-divergences studied in this work as well as the specific choices of activation function $g_f$ given by \citet{nowozin2016f} and utilized in \citet{ke2019imitation}. Also shown are the convex conjugates, inverse convex conjugates, and their respective domains.}
\label{tab:fdiv}
\end{table*}

\subsection{$f$-divergences and Imitation Learning}
\label{sec:fvim}

The GAIL and GAIFO approaches engage in an adversarial game where the discriminator estimates the divergence between state-action or state transition distributions according to the Jensen-Shannon divergence~\citep{goodfellow2014generative}. In this work, our focus is on a more general class of divergences, that includes the Jensen-Shannon divergence, known as Ali-Silvey distances or $f$-divergences~\citep{ali1966general,csiszar2004information}. For two distributions $P$ and $Q$ with support over a domain $\mathcal{X}$ and corresponding continuous densities $p$ and $q$, we have the $f$-divergence between them according to:
\begin{equation}
    D_f(P || Q) = \int\limits_\mathcal{X} q(x) f(\frac{p(x)}{q(x)}) dx
    \label{eq:fdiv}
\end{equation}

where $f:\mathbb{R}_+ \rightarrow \mathbb{R}$ is a convex, lower-semicontinuous function such that $f(1) = 0$. As illustrated in Table \ref{tab:fdiv}, different choices of function $f$ yield well-known divergences between probability distributions. 
\citet{nowozin2016f} appeal to the variational lower bound on $f$-divergences derived by \citet{nguyen2010estimating} (see Appendix for a full derivation) to extend GANs to arbitrary $f$-divergences, or $f$-GANs. Specifically, the two distributions of interest are the real data distribution $P$ and a synthetic distribution represented by a generative model $Q_\theta$ with parameters $\theta$. The variational function is also parameterized as $T_\omega$ acting as the discriminator. This gives rise to the VDM principle which defines the $f$-GAN objective
\begin{equation}
    \min\limits_\theta \max\limits_\omega \mathbb{E}_{x \sim P}[T_\omega(x)] - \mathbb{E}_{x \sim Q_\theta}[f^*(T_\omega(x))]
    \label{eq:fgan_obj}
\end{equation}

\citet{nowozin2016f} represent the variational function as $T_\omega(x) = g_f(V_\omega(x))$ such that $V_\omega(x):\mathcal{X} \rightarrow \mathbb{R}$ represents the unconstrained discriminator network while $g_f:\mathbb{R} \rightarrow \text{dom}_{f^*}$ is an activation function chosen in accordance with the $f$-divergence being optimized. Table \ref{tab:fdiv} includes the ``somewhat arbitrary'' but effective choices for $g_f$ suggested by \citet{nowozin2016f} and we refer readers to their excellent work for more details and properties of $f$-divergences and $f$-GANs. 

Recently, \citet{ke2019imitation} have formalized the generalization from GAN to $f$-GAN for the traditional IL problem. They offer the $f$-Variational Imitation ($f$-VIM) framework for the specific case of estimating and then minimizing the divergence between state-action distributions induced by expert and imitation policies:
\begin{multline}
    \min\limits_\theta \max\limits_\omega \mathbb{E}_{(s,a) \sim \rho^{\pie}}[g_f(V_\omega(s,a))] \\ - \mathbb{E}_{(s,a) \sim \rho^{\pi_\theta}}[f^*(g_f(V_\omega(s, a)))]
    \label{eq:fvim_obj}
\end{multline}

where $V_\omega: \mathcal{S} \times \mathcal{A} \rightarrow \mathbb{R}$ denotes the discriminator network that will supply per-timestep rewards during the outer policy optimization which itself is carried out over policy parameters $\theta$ via policy-gradient reinforcement learning~\citep{sutton2000policy}. In particular, the per-timestep rewards provided to the agent are given according to $r(s,a,s') = f^*(g_f(V_\omega(s,a)))$.

While \citet{ke2019imitation} do an excellent job of motivating the use of $f$-divergences for IL (by formalizing the relationship between divergences over trajectory distributions vs. state-action distributions) and connecting $f$-VIM to existing imitation-learning algorithms, their experiments focus on smaller problems to study the mode-seeking/mode-covering aspects of different $f$-divergences and the implications of such behavior depending on the multimodality of the expert trajectory distribution. Meanwhile, in the course of attempting to apply $f$-VIM to large-scale imitation problems, we empirically observe numerical instabilities stemming from function approximation, demanding a reformulation of the framework. Concurrently, the framework of \citet{kamyar2019adiv} also connects $f$-divergences and standard IL while evaluating on higher-dimensional MuJoCo tasks. We note that their IL evaluation largely focuses on the forward and reverse KL divergences while the implementation relies on carefully tuned logit thresholding and gradient norm clipping for the discriminator; in contrast, this work explores several other choices of $f$-divergence while our reparameterization directly addresses the numerical instability issues without the need for further heuristics to stabilize policy optimization.



\subsection{Reparameterizing $f$-VIM}
\label{sec:fvim_refac}

In their presentation of the $f$-VIM framework, \citet{ke2019imitation} retain the choices for activation function $g_f$ introduced by \citet{nowozin2016f} for $f$-GANs. Recall that these choices of $g_f$ play a critical role in defining the reward function optimized by the imitation policy on each iteration of $f$-VIM, $r(s,a,s') = f^*(g_f(V_\omega(s,a)))$. It is well known in the reinforcement-learning literature that the nature of the rewards provided to an agent have strong implications on learning success and efficiency~\citep{Ng1999PolicyIU,Singh2010IntrinsicallyMR}. While the activation choices made for $f$-GANs are suitable given that both optimization problems are carried out by backpropagation, we assert that special care must be taken when specifying these activations (and implicitly, the reward function) for imitation-learning algorithms. A combination of convex conjugate and activation function could induce a reward function that engenders numerical instability or a simply challenging reward landscape, depending on the underlying policy-gradient algorithm utilized~\citep{henderson2018deep}. Empirically, we found that the particular activation choices for the KL and reverse KL divergences shown in Table \ref{tab:fdiv} (linear and exponential, respectively) produced imitation-learning algorithms that, in all of our evaluation environments, failed to complete execution due to numerical instabilities caused by exploding policy gradients (avoided by \citet{kamyar2019adiv} through discriminator logit clipping that, implicitly, acts as reward clipping). In the case of the Total Variation distance, the corresponding $f$-GAN activation for the variational function is a $\tanh$, requiring a learning agent to traverse a reward interval of $[-1,1]$ by crossing an intermediate region with reward signals centered around $0$. 

To refactor the $f$-VIM framework so that it more clearly exposes the choice of reward function to practictioners and shifts the issues of reward scale away from the imitation policy, we propose uniformly applying an activation function $g_f(v) = f^{*-1}(r(v))$ where $f^{*-1}(t)$ denotes the inverse of the convex conjugate (see Table \ref{tab:fdiv}). Here $r$ is effectively a free parameter that can be set according to one of the many heuristics used throughout the field of deep reinforcement learning for maintaining a reasonable reward scale~\citep{Mnih2015HumanlevelCT,Mnih2016AsynchronousMF,henderson2018deep} so long as it obeys the domain of the inverse conjugate $\text{dom}_{f^{*-1}}$. In selecting $g_f$ accordingly, the reparameterized saddle-point optimization for $f$-VIM becomes
\begin{multline}
    \min\limits_\theta \max\limits_\omega \mathbb{E}_{(s,a) \sim \rho^{\pie}}[f^{*-1}(r(V_\omega(s,a)))] \\ - \mathbb{E}_{(s,a) \sim \rho^{\pi_\theta}}[r(V_\omega(s, a))]
    \label{eq:fvim_refac}
\end{multline}

where the per-timestep rewards used during policy optimization are given by $r(s,a,s') = r(V_\omega(s, a))$. In applying this choice, we shift the undesirable scale of the latter term in VDM towards the discriminator, expecting it to be indifferent since training is done by backpropagation. As one potential instantiation, we consider $r(u) = \sigma(u)$ where $\sigma(\cdot)$ denotes the sigmoid function leading to bounded rewards in the interval $[0,1]$ that conveniently adhere to $\text{dom}_{f^{*-1}}$ for almost all of the $f$-divergences examined in this work\footnote{For Total Variation distance, we use $r(u) = \frac{1}{2}\sigma(u)$ to remain within $\text{dom}_{f^{*-1}}$.}. In the Appendix, we discuss the empirically less-successful alternative of swapping the distribution positions in Equation \ref{eq:fgan_obj}. In Section \ref{sec:exps}, we evaluate IL algorithms with this choice against those using $f$-VIM with the original $f$-GAN activations; we find that, without regard for the scale of rewards and the underlying reinforcement-learning problem being solved, the $f$-GAN activation choices either produce degenerate solutions or completely fail to produce an imitation policy altogether.



\subsection{$f$-divergences and Imitation from Observation}
\label{sec:fvimo}

Applying the variational lower bound of \citet{nguyen2010estimating} and the corresponding $f$-GAN extension, we can now present our Variational Imitation from Observation ($f$-VIMO) extension for a general family of ILO algorithms that leverage the VDM principle in the underlying saddle-point optimization. Since optimization of the generator will continue to be carried out by policy-gradient reinforcement learning, we adhere to our reparameterization of the $f$-VIM framework and present the $f$-VIMO objective as:
\begin{multline}
    \min\limits_\theta \max\limits_\omega \mathbb{E}_{(s,s') \sim \rho^{\pie}}[f^{*-1}(r(V_\omega(s,s')))] \\ - \mathbb{E}_{(s,s') \sim \rho^{\pi_\theta}}[r(V_\omega(s, s'))]
    \label{eq:fvimo}
\end{multline}

with the per-timestep rewards given according to $r(s,a,s') = r(V_\omega(s, s'))$. We present the full approach as Algorithm \ref{alg:fvimo}. Just as in Section \ref{sec:fvim_refac}, we again call attention to Line 5 where the discriminator outputs (acting as individual reward signals) scale the policy gradient, unlike the more conventional discriminator optimization of Line 4 by backpropagation; this key difference is the primary motivator for our specific reparameterization of the $f$-VIM framework. Just as in the previous section, we take $r(u) = \sigma(u)$ as a particularly convenient choice of activation given its agreement to the inverse conjugate domains $\text{dom}_{f^{*-1}}$ for many choices of $f$-divergence and we employ this instantiation throughout all of our experiments. We leave the examination of alternative choices for $r$ to future work.

\begin{algorithm}[tb]
\begin{algorithmic}
\STATE \textsc{Input:} Dataset of expert trajectories $\mathcal{D}$, initial policy and discriminator parameters $\theta_0$ and $\omega_0$, number of iterations $N$, discount factor $\gamma$
\FOR{$i=0,1,\ldots,N$}
\STATE Sample trajectories from current imitation policy $\tau_i \sim \pi_{\theta_i}$
\STATE $\omega_{i+1} = \omega_i + \nabla_\omega \Big( \mathbb{E}_{(s,s') \sim \mathcal{D}}[f^{*-1}(r(V_\omega(s,s')))] - \mathbb{E}_{(s,s') \sim \tau_i}[r(V_\omega(s, s'))] \Big)$
\STATE Update $\theta_i$ to $\theta_{i+1}$ via a policy-gradient update with rewards given by $r(V_\omega(s, s'))$:
\STATE $G_i = \mathbb{E}_{\tau_i}[\sum\limits_{t=1}^{\infty}\gamma^{t-1}r(V_\omega(s_{t-1}, s_{t})) | s_0=s, s_1 = s']$
\STATE $\theta_{i+1} = \theta_i + \mathbb{E}_{(s,a,s') \sim \tau_i}\Big[\nabla_\theta \log(\pi_{\theta_i}(a|s))G_i\Big]$
\ENDFOR
\end{algorithmic}
\caption{$f$-VIMO}
\label{alg:fvimo}
\end{algorithm}

\subsection{Discriminator Regularization}
\label{sec:discr_reg}

The refactored version of $f$-VIM presented in Section \ref{sec:fvim_refac} is fundamentally addressing instability issues that may occur on the generator side of adversarial training; in our experiments, we also examine the utility of regularizing the discriminator side of the optimization for improved stability. Following from a line of work examining the underlying mathematical properties of GAN optimization~\citep{roth2017stabilizing,roth2018adversarially,mescheder2018training}, we opt for the simple gradient-based regularization of \citet{mescheder2018training} which (for $f$-VIMO) augments the discriminator loss with the following regularization term:
\begin{equation}
    R(\omega) = \frac{\psi}{2} \mathbb{E}_{(s,s') \sim \rho^{\pie}}[||\nabla_\omega f^{*-1}(r(V_\omega(s,s'))) ||^2]
    \label{eq:discr_reg}
\end{equation}

where $\psi$ is a hyperparameter controlling the strength of the regularization. The form of this specific penalty follows from the analysis of \citet{roth2017stabilizing}; intuitively, its purpose is to disincentivize the discriminator from producing a non-zero gradient that shifts away from the Nash equilibrium of the minimax optimization  when presented with a generator that perfectly matches the true data distribution. While originally developed for traditional GANs and shown to empirically exhibit stronger convergence properties over Wasserstein GANs~\citep{Gulrajani2017ImprovedTO}, this effect is still desirable for the adversarial IL setting where the reward function (discriminator) used for optimizing the imitation policy should stop changing once the expert state-transition distribution has been matched. In practice, we compare $f$-VIM and $f$-VIMO both with and without the use of this regularization term and find that $R(\omega)$ can improve the stability and convergence of $f$-VIMO across almost all domains.




\begin{figure*}[h!]
    \centering
    \includegraphics[width=0.75\linewidth]{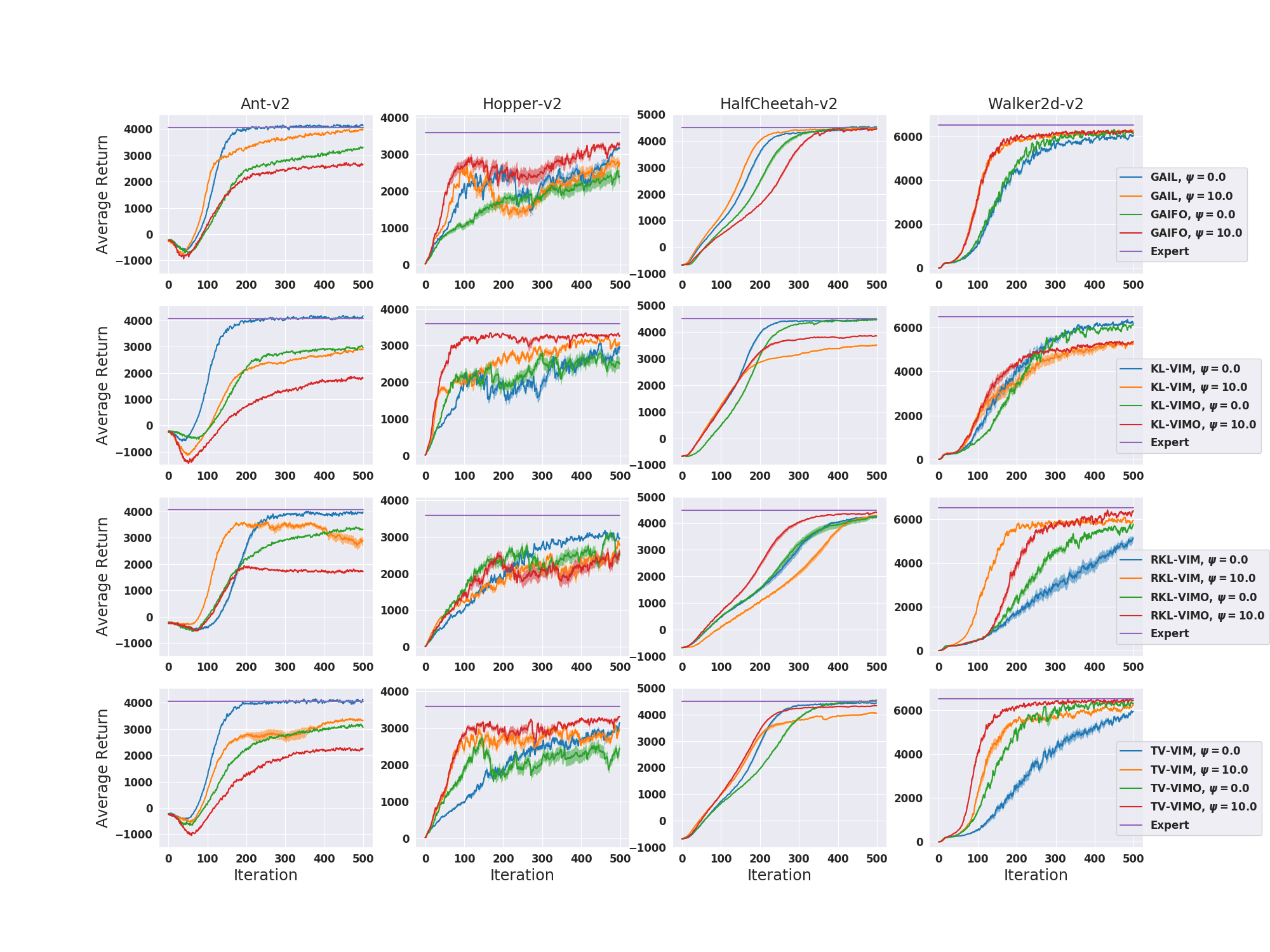}
    \caption{Comparing $f$-VIM and $f$-VIMO across four MuJoCo environments with $f$-divergences: GAN, Kullback-Leibler (KL), reverse KL (RKL), and Total Variation distance (TV). All algorithms were provided $20$ expert demonstrations and we also examine effect of discriminator regularization (Equation \ref{eq:discr_reg}) with $\psi=10$ per \citet{mescheder2018training}.}
    \label{fig:fvim_vimo_reg}
\end{figure*}

\section{Experiments}
\label{sec:exps}

We examine four instantiations of the $f$-VIM and $f$-VIMO frameworks (as presented in Sections \ref{sec:fvim_refac} and \ref{sec:fvimo}) corresponding to imitation algorithms with the following choices of $f$-divergence: GAN, Kullback-Leibler, reverse KL, and Total Variation\footnote{Code available at \url{https://github.com/DilipA/fVIMO}}. We conduct our evaluation across four MuJoCo environments~\citep{todorov2012mujoco} of varying difficulty: Ant, Hopper, HalfCheetah, and Walker (please see the Appendix for more details on individual environments). 

The core questions we seek to answer through our empirical results are as follows:
\begin{enumerate}
    \item What are the implications of the choice of activation for the variational function in $f$-VIM on imitation policy performance?
    \item Do $f$-divergences act as a meaningful axis of variation for IL and ILO algorithms?
    \item What is the impact of discriminator regularization on the stability and convergence properties of $f$-VIM/$f$-VIMO?
    \item How does the impact of different $f$-divergences vary with the amount of expert demonstration data provided?
\end{enumerate}

Due to space constraints, we defer results pertaining to our first experimental question to the Appendix where we illustrate the importance of activation function (reward function) to the overall adversarial imitation problem. In order to obtain reportable results for KL and RKL divergences, we employ gradient norm clipping, finding that laborious tuning of such heuristics may still lead to suboptimal imitation policies.

To answer the second and third questions above, we report the average total reward achieved by the imitation policy throughout the course of learning with rewards as defined by the corresponding OpenAI Gym environment~\citep{Brockman2016OpenAIG}. Shading in all plots denote $95\%$ confidence intervals computed over 10 random trials with 10 random seeds. Expert demonstration datasets of $50$ trajectories were collected from agents trained via Proximal Policy Optimization (PPO)~\citep{Schulman2017ProximalPO}; $20$ expert demonstrations were randomly subsampled at the start of learning and held fixed for the duration of the algorithm. We also utilize PPO as the underlying reinforcement-learning algorithm for training the imitation policy with a clipping parameter of $0.2$, advantage normalization, entropy regularization coefficient $0.001$, and the Adam optimizer~\citep{Kingma2014AdamAM}. Just as in \citet{ho2016generative} we use a discount factor of $\gamma = 0.995$ and apply Generalized Advantage Estimation~\citep{schulman2015high} with parameter $\lambda = 0.97$. We run both $f$-VIM and $f$-VIMO for a total of $500$ iterations, collecting $50000$ environment samples per iteration. The policy and discriminator architectures are identically two separate multi-layer perceptrons each with two hidden layers of $100$ units separated by $\tanh$ nonlinearities. A grid search was used for determining the initial learning rate, number of PPO epochs, and number of epochs used for discriminator training (we refer readers to the Appendix for more details) and we report results for the best hyperparameter settings.

To address our final question, we take the best hyperparameter settings recovered when given $20$ expert demonstrations and re-run all algorithms with $\{1,5,10,15\}$ expert demonstrations that are randomly sampled at the start of each random trial and held fixed for the duration of the algorithm. We then record the average return of the final imitation policy for each level of expert demonstration. Due to space constraints, all additional empirical results are presented in the Appendix.

\begin{figure*}[h!]
    \centering
    \includegraphics[width=0.75\linewidth]{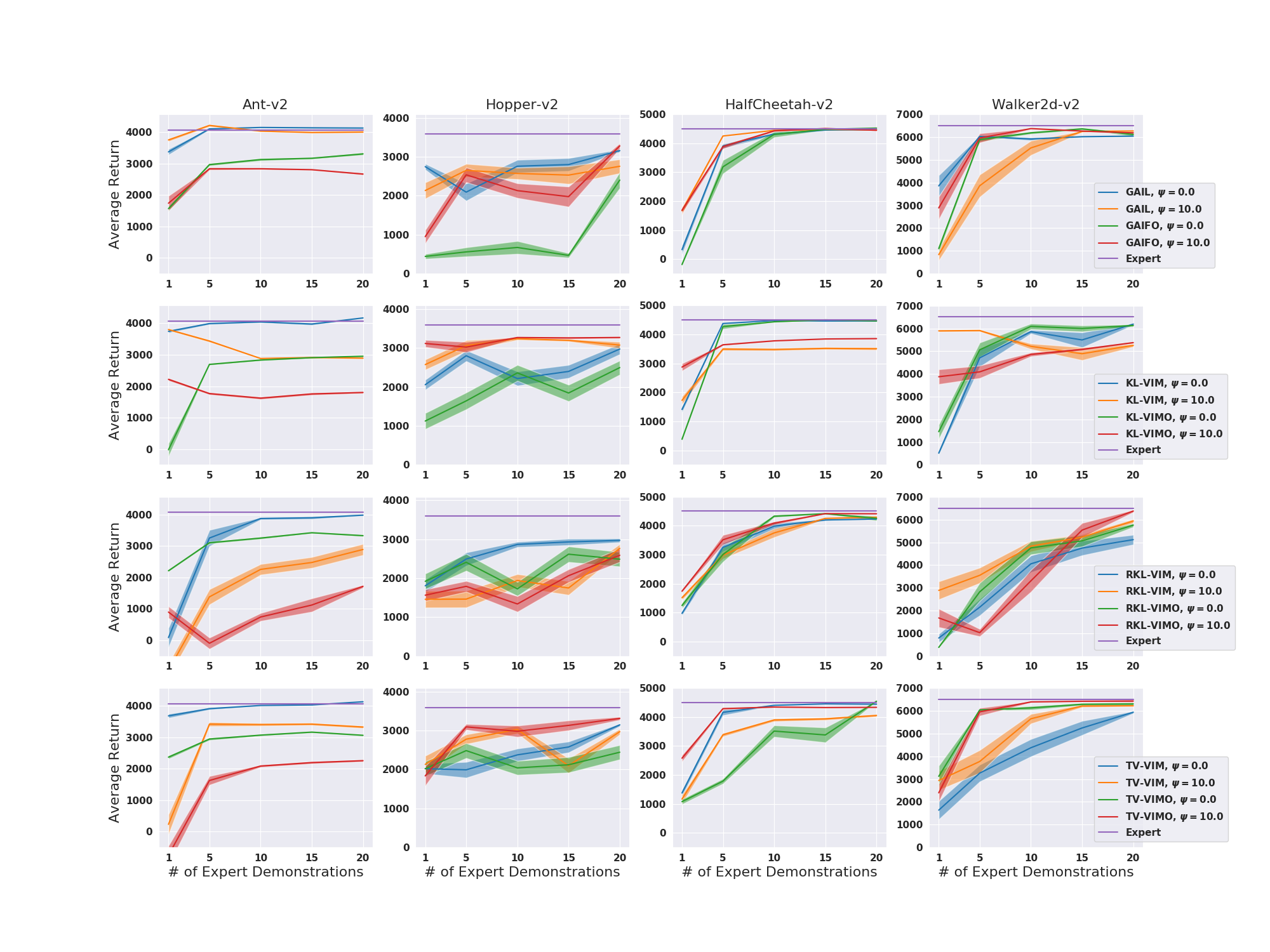}
    \caption{Comparing $f$-VIM and $f$-VIMO according to final imitation policy performances under varying amounts of expert demonstration data.}
    \label{fig:fvim_vimo_sc}
\end{figure*}

\section{Results \& Discussion}



We present results in Figure \ref{fig:fvim_vimo_reg} to assess the utility of varying the choice of divergence in $f$-VIM and $f$-VIMO across each domain. In considering the impact of $f$-divergence choice, we find that most of the domains must be examined in isolation to observe a particular subset of $f$-divergences that stand out. In the IL setting, we find that varying the choice of $f$-divergence can yield different learning curves but, ultimately, produce near-optimal (if not optimal) imitation policies across all domains. In contrast, we find meaningful choices of $f$-divergence in the ILO setting including $\{$KL, TV$\}$ for Hopper, RKL for HalfCheetah, and $\{$GAN, TV$\}$ for Walker. We note that the use of discriminator regularization per \citet{mescheder2018training} is crucial to achieving these performance gains, whereas the regularization generally fails to help performance in the IL setting. This finding is supportive of the logical intuition that ILO poses a fundamentally more-challenging problem than standard IL. Our success with the (forward) KL divergence runs contrary to the findings of \citet{kamyar2019adiv} who assert that Equation \ref{eq:fvim_obj} cannot be used under the choice of the KL divergence. We posit that this occurs by violating their assumption of an optimal variational function (\citet{kamyar2019adiv} -- Appendix E), through a combination of our reparameterization as well as running a fixed number of discriminator updates in the inner-loop maximization.

As a negative result, we find that the Ant domain (the most difficult environment with $\mathcal{S} \subset \mathbb{R}^{111}$ and $\mathcal{A} \subset \mathbb{R}^{8}$) still poses a challenge for ILO algorithms across the board. More specifically, we observe that discriminator regularization hurts learning in both the IL and ILO settings. While the choice of RKL does manage to produce a marginal improvement over GAIFO, the gap between existing state-of-the-art and expert performance remains unchanged. It is an open challenge for future work to either identify the techniques needed to achieve optimal imitation policies from observations only or characterize a fundamental performance gap when faced with sufficiently large observation spaces. 

In Figure \ref{fig:fvim_vimo_sc}, we vary the total number of expert demonstrations available during learning and observe that certain choices of $f$-divergences can be more robust in the face of less expert data, both in the IL and ILO settings. We find that KL-VIM and TV-VIM are slightly more performant than GAIL when only provided with a single expert demonstration. Notably, in each domain we see that certain choices of divergence for $f$-VIMO do a better job of residing close to their $f$-VIM counterparts suggesting that future improvements may come from examining $f$-divergences in the small-data regime. This idea is further exemplified when accounting for results collected while using discriminator regularization~\citep{mescheder2018training}. We refer readers to the Appendix for the associated learning curves.

Our work leaves many open directions for future work to close the performance gap between student and expert policies in the ILO setting. While we found the sigmoid function to be a suitable instantiation of our framework, exploring alternative choices of variational function activations could prove useful in synthesizing performant ILO algorithms. Alternative choices of $f$-divergences could lead to more substantial improvements than the choices we examine in this paper. Moreover, while this work has a direct focus on $f$-divergences, Integral Probability Metrics (IPMs)~\citep{muller_1997,gretton2012kernel} represent a distinct but well-established family of divergences between probability distributions. The success of Total Variation distance in our experiments, which doubles as both a $f$-divergence and IPM~\citep{sriperumbudur2009integral}, is suggestive of future work building IPM-based ILO algorithms~\citep{sun2019provably}.

\section{Conclusion}

In this work, we present a general framework for IL and ILO under arbitrary choices of $f$-divergence. We empirically validate a single instantiation of our framework across multiple $f$-divergences demonstrating that, unlike its predecessors, our reparameterization can scale to complex tasks without the need for laborious tuning of clipping parameters.





\bibliography{references}
\bibliographystyle{icml2020}

\onecolumn

\icmltitle{Reparameterized Variational Divergence Minimization for Stable Imitation Appendix}



\icmlsetsymbol{equal}{*}

\begin{icmlauthorlist}
\icmlauthor{Dilip Arumugam}{st}
\icmlauthor{Debadeepta Dey}{msr}
\icmlauthor{Alekh Agarwal}{msr}
\icmlauthor{Asli Celikyilmaz}{msr}
\icmlauthor{Elnaz Nouri}{msr}
\icmlauthor{Bill Dolan}{msr}
\end{icmlauthorlist}

\icmlaffiliation{st}{Department of Computer Science, Stanford University, Stanford, California, USA}
\icmlaffiliation{msr}{Microsoft Research, Redmond, Washington, USA}

\icmlcorrespondingauthor{Dilip Arumugam}{dilip@cs.stanford.edu}
\icmlcorrespondingauthor{Debadeepta Dey}{dedey@microsoft.com}

\icmlkeywords{Machine Learning, ICML}

\vskip 0.3in




\section{Related Work}

\subsection{Learning from Demonstration}

Our work broadly falls within the category of Learning from Demonstration (LfD)~\citep{Schaal1997learning,Atkeson1997RobotLF,Argall2009ASO}, where an agent must leverage demonstration data (typically provided as trajectories, each consisting of expert state-action pairs) to produce an imitation policy that correctly captures the demonstrated behavior. Within the context of LfD, a finer distinction can be made between behavioral cloning (BC)~\citep{bain1999framework,pomerleau1989alvinn} and inverse reinforcement learning (IRL)~\citep{ng2000algorithms,Abbeel2004ApprenticeshipLV,Syed2007AGA,ziebart2008maximum,finn2016guided,ho2016generative} approaches; BC approaches view the demonstration data as a standard dataset of input-output pairs and apply traditional supervised-learning techniques to recover an imitation policy. Alternatively, IRL-based methods synthesize an estimate of the reward function used to train the expert policy before subsequently applying a reinforcement-learning algorithm~\citep{sutton1998introduction,Abbeel2004ApprenticeshipLV} to recover the corresponding imitation policy. Although not a focus of this work, we also acknowledge the myriad of approaches that operate at the intersection of IL and reinforcement learning or augment reinforcement learning with IL~\citep{rajeswaran2017learning,hester2018deep,salimans2018learning,sun2018truncated,borsa2019observational,tirumala2019exploiting}. 

While BC approaches have been successful in some settings~\citep{niekum2015learning,Giusti2016AML,bojarski2016end}, they are also susceptible to failures stemming from covariate shift where minute errors in the actions of the imitation policy compound and force the agent into regions of the state space not captured in the original demonstration data. While some preventative measures for covariate shift do exist~\citep{laskey2017dart}, a more principled solution can be found in methods like DAgger~\citep{ross2011reduction} and its descendants~\citep{ross2014reinforcement,sun2017deeply,le2018hierarchical} that remedy covariate shift by querying an expert to provide on-policy action labels. It is worth noting, however, that these approaches are only feasible in settings that admit such online interaction with an expert~\citep{laskey2016shiv} and, even then, failure modes leading to poor imitation policies do exist~\citep{laskey2017comparing}.

The algorithms presented in this work fall in with IRL-based approaches to IL. Early successes in this regime tend to rely on hand-engineered feature representations for success~\citep{Abbeel2004ApprenticeshipLV,ziebart2008maximum,levine2011nonlinear}. Only in recent years, with the aid of deep neural networks, has there been a surge in the number of approaches that are capable of scaling to the raw, high-dimensional observations found in real-world control problems~\citep{finn2016guided,ho2016generative,duan2017one,li2017infogail,fu2018learning,kim2018imitation}. Our work focuses attention exclusively on adversarial methods for their widespread effectiveness across a range of imitation tasks without requiring interactive experts~\citep{ho2016generative,li2017infogail,fu2018learning,Kostrikov2018DiscriminatorActorCriticAS}; at the heart of these methods is the Generative Adversarial Imitation Learning (GAIL)~\citep{ho2016generative} approach which produces high-fidelity imitation policies and achieves state-of-the-art results across numerous continuous-control benchmarks by leveraging the expressive power of Generative Adversarial Networks (GANs)~\citep{goodfellow2014generative} for modeling complex distributions over a high-dimensional support. From an IRL perspective, GAIL can be viewed as iteratively optimizing a parameterized reward function (discriminator) that, when used to optimize an imitation policy (generator) via policy-gradient reinforcement learning~\citep{sutton2000policy}, allows the agent to shift its own behavior closer to that of the expert. From the perspective of GANs, this is achieved by discriminating between the respective distributions over state-action pairs visited by the imitation and expert policies before training a generator to fool the discriminator and induce a state-action visitation distribution similar to that of the expert.

While a large body of prior work exists for IL, numerous recent works have drawn attention to the more challenging problem of imitation learning from observation~\citep{Sermanet2017TimeContrastiveNS,Liu2018ImitationFO,Goo2018OneShotLO,Kimura2018InternalMF,torabi2018behavioral,torabi2018generative,edwards2019imitating,sun2019provably}. In an effort to more closely resemble observational learning in humans and leverage the wealth of publicly-available, observation-only data sources, the ILfO problem considers learning from expert demonstration data where no expert action labels are provided. Many early approaches to ILfO use expert observation sequences to learn a semantic embedding space so that distances between observation sequences of the imitation and expert policies can serve as a cost signal to be minimized via reinforcement learning~\citep{Gupta2017LearningIF,Sermanet2017TimeContrastiveNS,Dwibedi2018LearningAR,Liu2018ImitationFO}. In contrast, \citet{torabi2018behavioral} introduce Behavioral Cloning from Observation (BCO) which leverages state-action trajectories collected under a random policy to train an inverse dynamics model for inferring the action responsible for a transition between two input states (assuming the two represent a state and next-state pair). With this inverse model in hand, the observation-only demonstration data can be converted into the more traditional dataset of state-action pairs over which standard BC can be applied. Recognizing the previously discussed limitations of BC approaches, \citet{torabi2018generative} introduce the natural GAIL counterpart for ILfO, Generative Adversarial Imitation from Observation (GAIFO); GAIFO is identical to GAIL except the distributions under consideration in the adversarial game are over state transitions (state and next-state pairs), as opposed to state-action pairs requiring expert action labels. While \citet{torabi2018generative} offer empirical results for continuous-control tasks with low-dimensional features as well as raw image observations, GAIFO falls short of expert performance in both settings leaving an open challenge for scalable ILfO algorithms that achieve expert performance across a wide spectrum of tasks. A central question of this work is to explore how alternative formulations of the GAN objective that underlies GAIFO might yield superior ILfO algorithms. For a more in-depth survey of ILfO approaches, we refer readers to \citet{Torabi2019RecentAI}.

\subsection{Generative Adversarial Networks}

 With a focus on generative-adversarial methods for IL, this work leverages several related ideas in the GAN literature for offering alternative formulations as well as improving understanding of their underlying mathematical foundations~\citep{li2015generative,dziugaite2015training,zhao2016energy,nowozin2016f,roth2017stabilizing,arjovsky2017wasserstein,Gulrajani2017ImprovedTO,roth2018adversarially,mescheder2018training}. Critical to the ideas presented in many of these previous works is an understanding that discriminator networks are estimating a divergence between two probability distributions of interest, usually taken to be the real data distribution and the fake or synthetic distribution represented by the generator. Formal characterizations of this divergence, either by Integral Probability Metrics (IPMs)~\citep{muller_1997,gretton2012kernel} or $f$-divergences~\citep{ali1966general,csiszar2004information,liese2006divergences}, yield different variations on the classic GAN formulation which is itself a slight variation on the Jensen-Shannon (JS) divergence~\citep{li2015generative,dziugaite2015training,zhao2016energy,nowozin2016f,arjovsky2017wasserstein,Gulrajani2017ImprovedTO}. Following from work by \citet{nowozin2016f} to generalize the GAN objective to arbitrary $f$-divergences, \citet{ke2019imitation} offer a generalization of GAIL to an arbitrary choice of $f$-divergence for quantifying the gap between the state-action visitation distributions of the imitation and expert policies; moreover, \citet{ke2019imitation} propose a unifying framework for IL, $f$-Variational IMitation ($f$-VIM), in which they highlight a correspondence between particular choices of $f$-divergences and existing IL algorithms (specifically BC $\Longleftrightarrow$ Kullback-Leibler (KL) divergence, DAgger $\Longleftrightarrow$ Total-Variation distance, and GAIL $\Longleftrightarrow$ JS-divergence~\footnote{The discriminator loss optimized in the original GAN formulation is $2 \cdot D_{JS} - \log(4)$ where $D_{JS}$ denotes the Jensen-Shannon divergence~\citep{goodfellow2014generative,nowozin2016f}.}). While \citet{ke2019imitation} focus on providing empirical results in smaller toy problems to better understand the interplay between $f$-divergence choice and the multimodality of the expert trajectory distribution, we provide an empirical evaluation of their $f$-VIM framework across a range of continous control tasks in the Mujoco domain~\citep{todorov2012mujoco}. Empirically, we find that some of the design choices $f$-VIM inherits from the original $f$-GAN work~\citep{nowozin2016f} are problematic when coupled with adversarial IL and training of the generator by policy-gradient reinforcement learning, instead of via direct backpropagation as in traditional GANs. Consequently, we refactor their framework to expose this point and provide one practical instantiation that works well empirically. We then go on to extend the $f$-VIM framework to the IFO problem ($f$-VIMO) and evaluate the resulting algorithms empirically against the state-of-the-art, GAIFO.  

\section{Experiment Details}
\label{sec:exp_details}

Here we provide details of the MuJoCo environments~\citep{todorov2012mujoco} used in our experiments as well as the details of the hyperparameter search conducted for all algorithms (IL and ILO) presented.

\subsection{MuJoCo Environments}

All environments have continuous observation and action spaces of varying dimensionality (as shown below). All algorithms evaluated in each environment were trained for a total of $500$ iterations, collecting $50,000$ environment transitions per iteration.

\begin{table}[H]
\centering
\begin{tabular}{ccc}
\toprule
    Task & Observation Space & Action Space \\ \hline
    Ant-v2 & $\mathbb{R}^{111}$ & $\mathbb{R}^{8}$\\
    Hopper-v2 & $\mathbb{R}^{11}$ & $\mathbb{R}^{3}$\\
    HalfCheetah-v2 & $\mathbb{R}^{17}$ & $\mathbb{R}^{6}$\\
    Walker2d-v2 & $\mathbb{R}^{17}$ & $\mathbb{R}^{6}$
\end{tabular}
\end{table}

\subsection{Hyperparameters}
Below we outline the full set of hyperparameters examined for all experiments presented in this work. We conducted a full grid search over $10$ random trials with $10$ random seeds and report results for the best hyperparameter setting. 

\begin{table}[H]
\centering
\begin{tabular}{ccc}
\toprule
    Hyperparameter & Values & \\ \hline
    Discriminator learning rate & $\{1\text{e}^{-4}, 1\text{e}^{-3}\}$\\
    PPO epochs & $\{5, 10\}$\\
    Discriminator epochs & $\{1, 5, 10\}$\\
\end{tabular}
\end{table}

Preliminary experiments were conducted to test smaller values for PPO epochs and policy learning rates before settling on the grid shown above.

\section{$f$-divergence Lower Bound}
\label{sec:fdiv_deriv}

In order to accommodate the tractable estimation of $f$-divergences when only provided samples from $P$ and $Q$, \citet{nguyen2010estimating} offer an approach for variational estimation of $f$-divergences. Central to their procedure is the use of the convex conjugate function or Fenchel conjugate~\citep{HiriartUrruty2004FundamentalsOC}, $f^*$, which exists for all convex, lower-semicontinuous functions $f$ and is defined as the following supremum:
\begin{align}
    f^*(t) &= \sup\limits_{u \in \text{dom}_{f}} \{ut - f(u)\}
    \label{eq:cconjugate}
\end{align}

Using the duality of the convex conjugate ($f^{**} = f$), \citet{nguyen2010estimating} represent $f(u) = \sup\limits_{t \in \text{dom}_{f^*}}\{tu - f^*(t)\}$ enabling a variational bound:
\begin{align}
    D_f(P||Q) &= \int\limits_\mathcal{X} q(x) \sup\limits_{t \in \text{dom}_{f^*}}\bigg\{t\frac{p(x)}{q(x)} - f^*(t)\bigg\} dx \nonumber \\ 
    &\geq \sup\limits_{T \in \mathcal{T}}(\int\limits_\mathcal{X} p(x)T(x)dx - \int\limits_\mathcal{X}q(x)f^*(T(x))dx) \nonumber \\ 
    &= \sup\limits_{T \in \mathcal{T}} (\mathbb{E}_{x \sim P}[T(x)] - \mathbb{E}_{x \sim Q}[f^*(T(x))])
\end{align}

where $\mathcal{T}$ is an arbitrary class of functions $T:\mathcal{X} \rightarrow \text{dom}_{f^*}$.

\newpage

\section{Additional Experiments}
\label{sec:more_exps}

\subsection{Stability for KL and RKL Divergences}

In the plots that follow, we analyze $f$-VIM using the KL and RKL divergences along with the original activation functions proposed by \citet{nowozin2016f} and re-used in \citet{ke2019imitation}. In all of our initial experiments, none of the trials completed due to numerical instability caused by exploding gradients. Consequently, to obtain reportable results, we follow suit with prior work on imitation learning with $f$-divergences~\citep{kamyar2019adiv} and employ gradient norm clipping to stabilize the optimization. 

For our results with the KL divergence, we note that gradient norm clipping, while bringing stability to the optimization, still fails to achieve the same degree of performance as our $f$-VIM-sigmoid. The concentration of imitation policy performance for higher clipping thresholds suggests that further increases in threshold would be unlikely to produce more favorable results.

In the case of RKL, we see a slightly different story where the stability of gradient clipping still fails to produce nontrivial imitation policies. In fact, several threshold values below are not reported due to continued instabilities (for higher threshold values) or due to negative performance values that would distort the view of the remaining results.  

\begin{figure*}[h!]
    \centering
    \includegraphics[width=.75\linewidth]{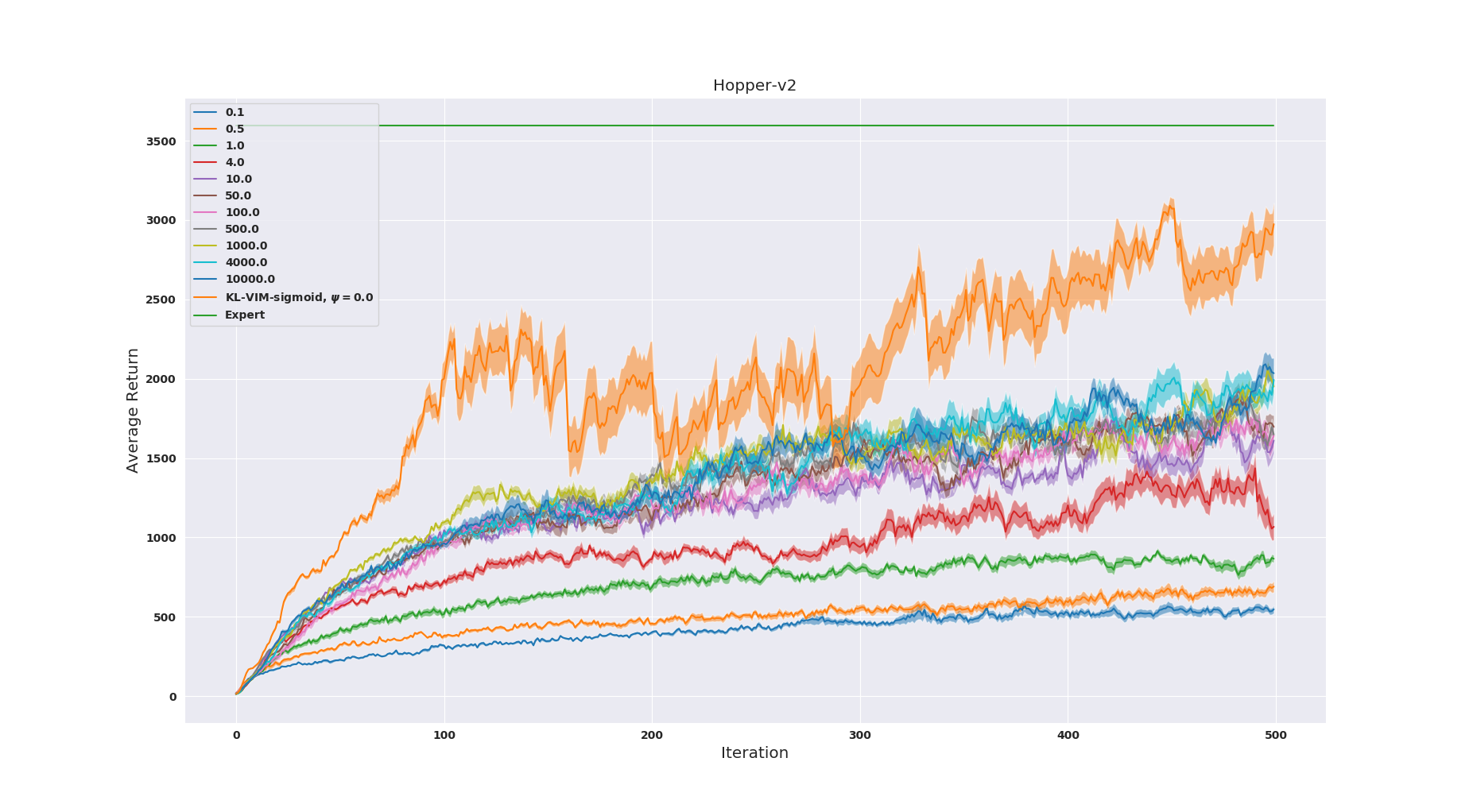}
    \caption{Examining stability of KL divergence in the Hopper environment, varying the threshold for gradient norm clipping.}
\end{figure*}

\newpage
\begin{figure*}[h!]
    \centering
    \includegraphics[width=.75\linewidth]{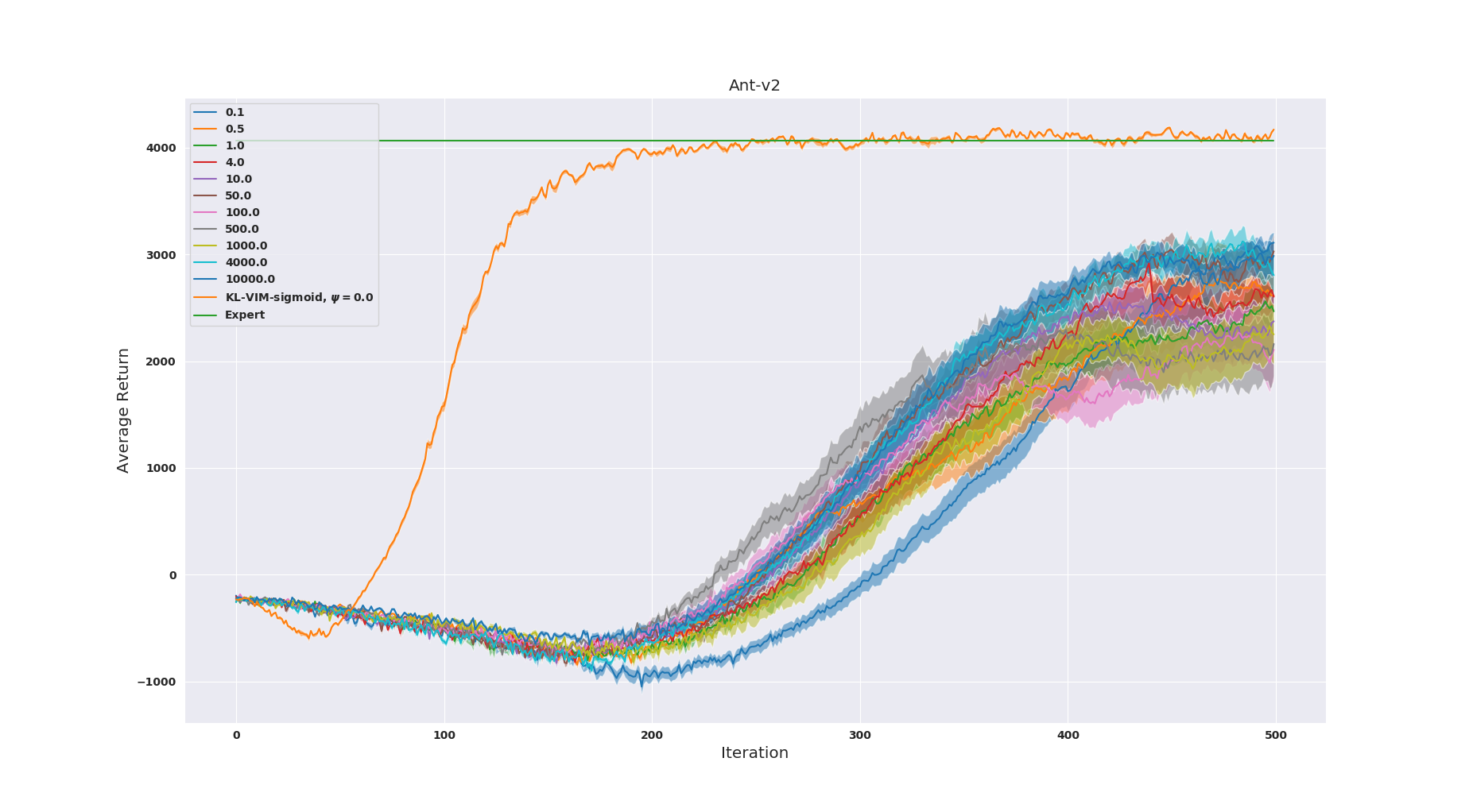}
    \caption{Examining stability of KL divergence in the Ant environment, varying the threshold for gradient norm clipping.}
\end{figure*}

\begin{figure*}[h!]
    \centering
    \includegraphics[width=.75\linewidth]{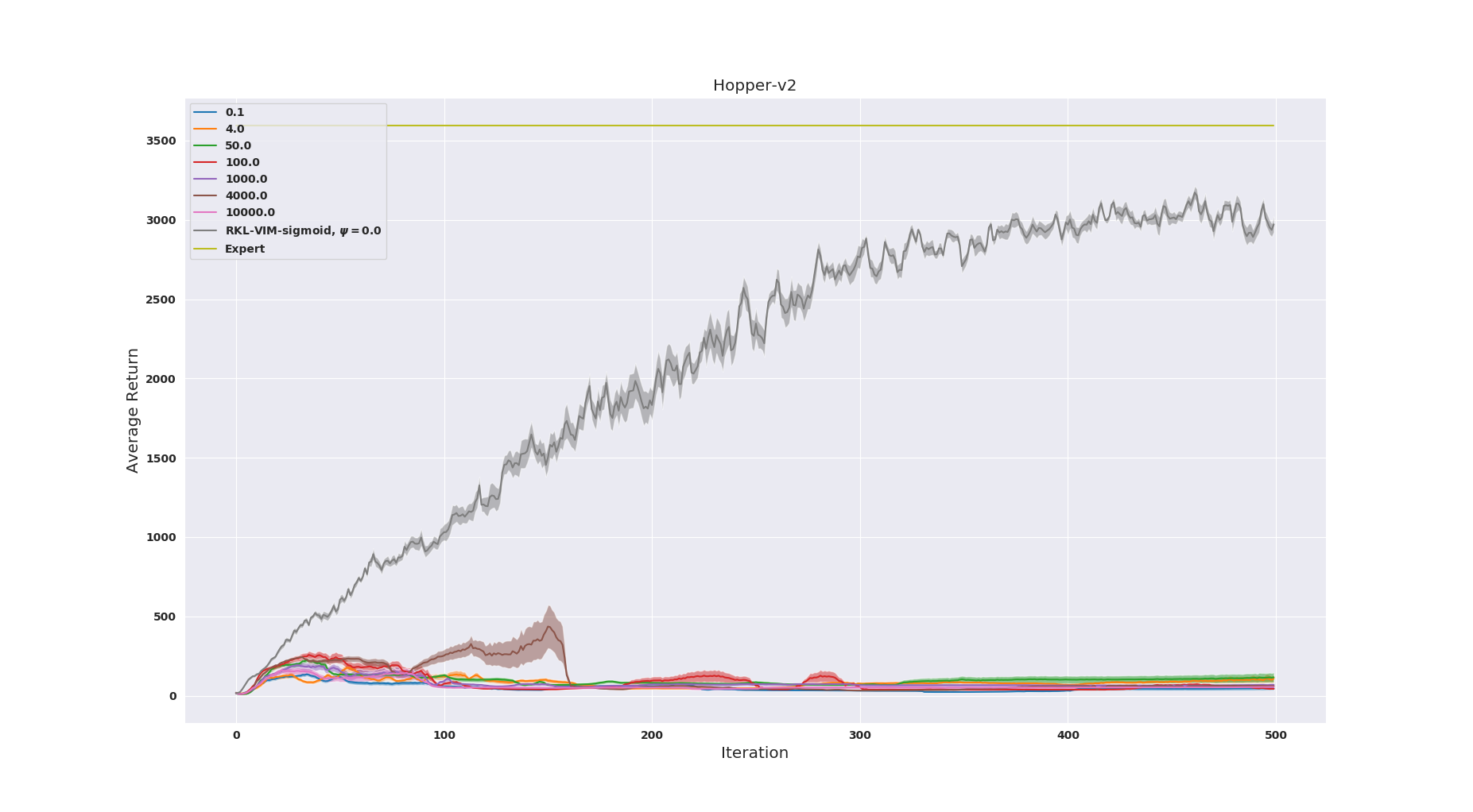}
    \caption{Examining stability of RKL divergence in the Hopper environment, varying the threshold for gradient norm clipping.}
\end{figure*}

\begin{figure*}[h!]
    \centering
    \includegraphics[width=.75\linewidth]{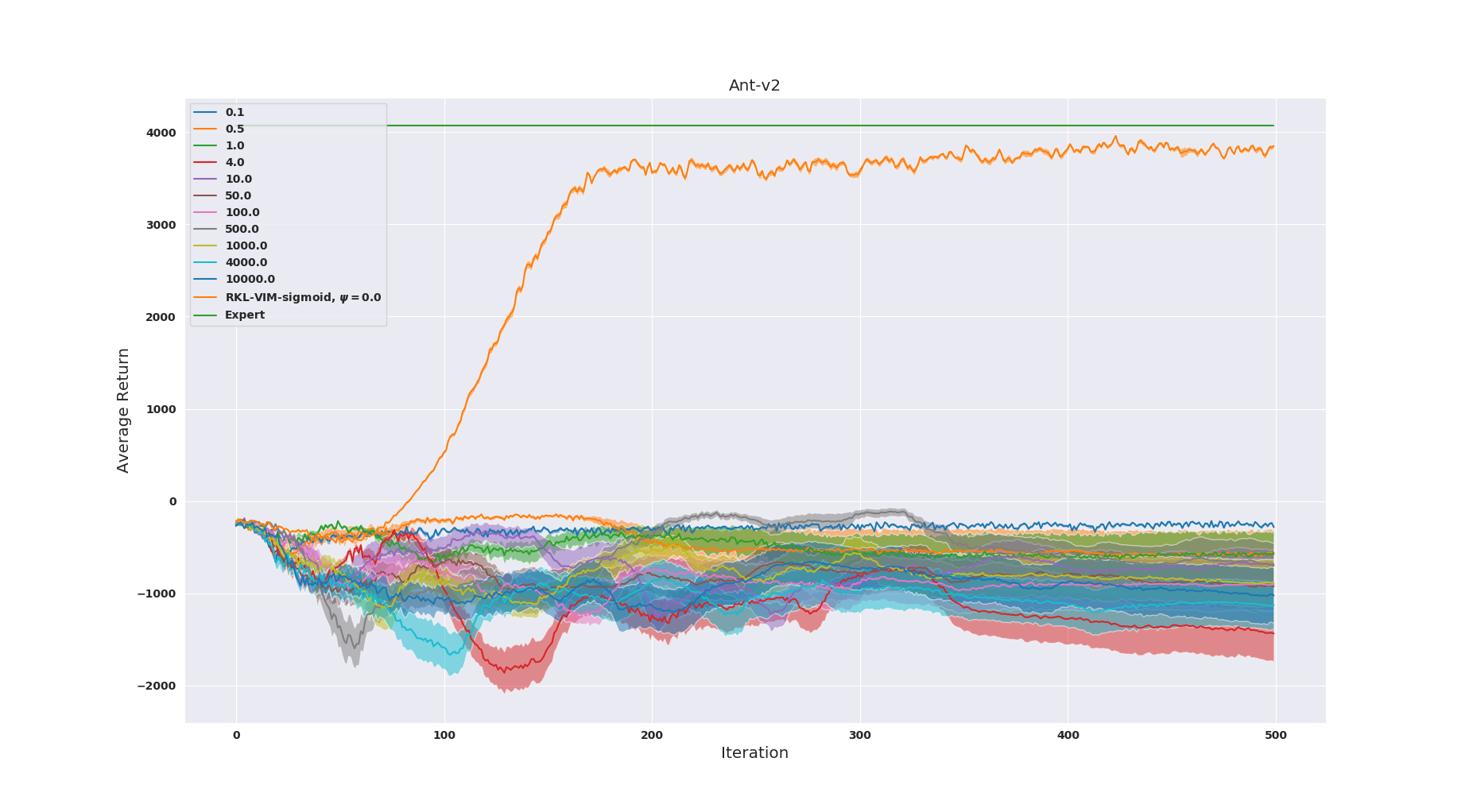}
    \caption{Examining stability of RKL divergence in the Ant environment, varying the threshold for gradient norm clipping.}
\end{figure*}
\newpage

\subsection{Unregularized $f$-VIM/VIMO}

\begin{figure*}[h!]
    \centering
    \includegraphics[width=.8\linewidth]{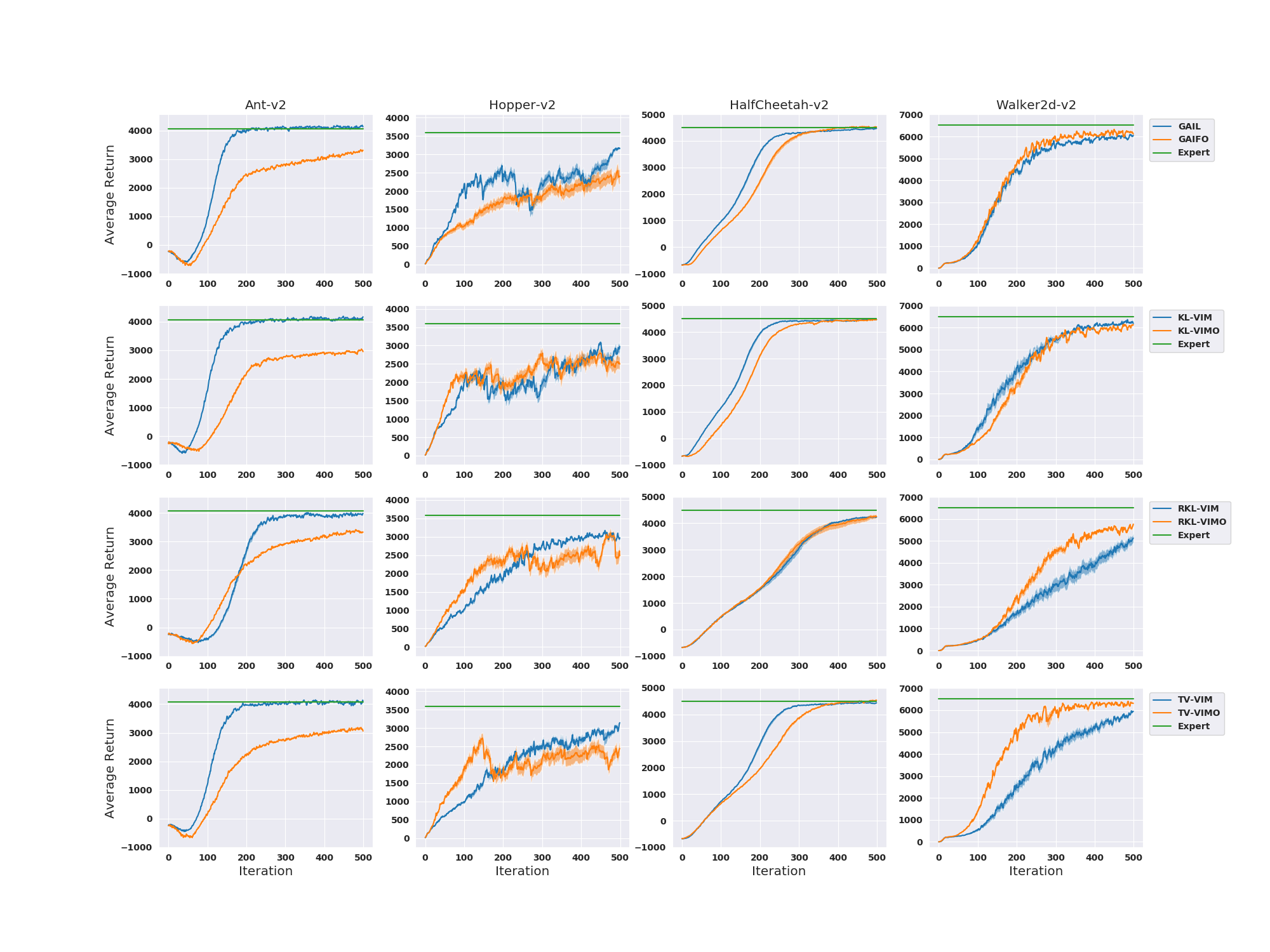}
    \caption{Comparing $f$-VIM and $f$-VIMO across four MuJoCo environments with $f$-divergences: GAN, Kullback-Leibler (KL), reverse KL (RKL), and Total Variation distance (TV).}
\end{figure*}

\newpage
\begin{figure*}[h!]
    \centering
    \includegraphics[width=.8\linewidth]{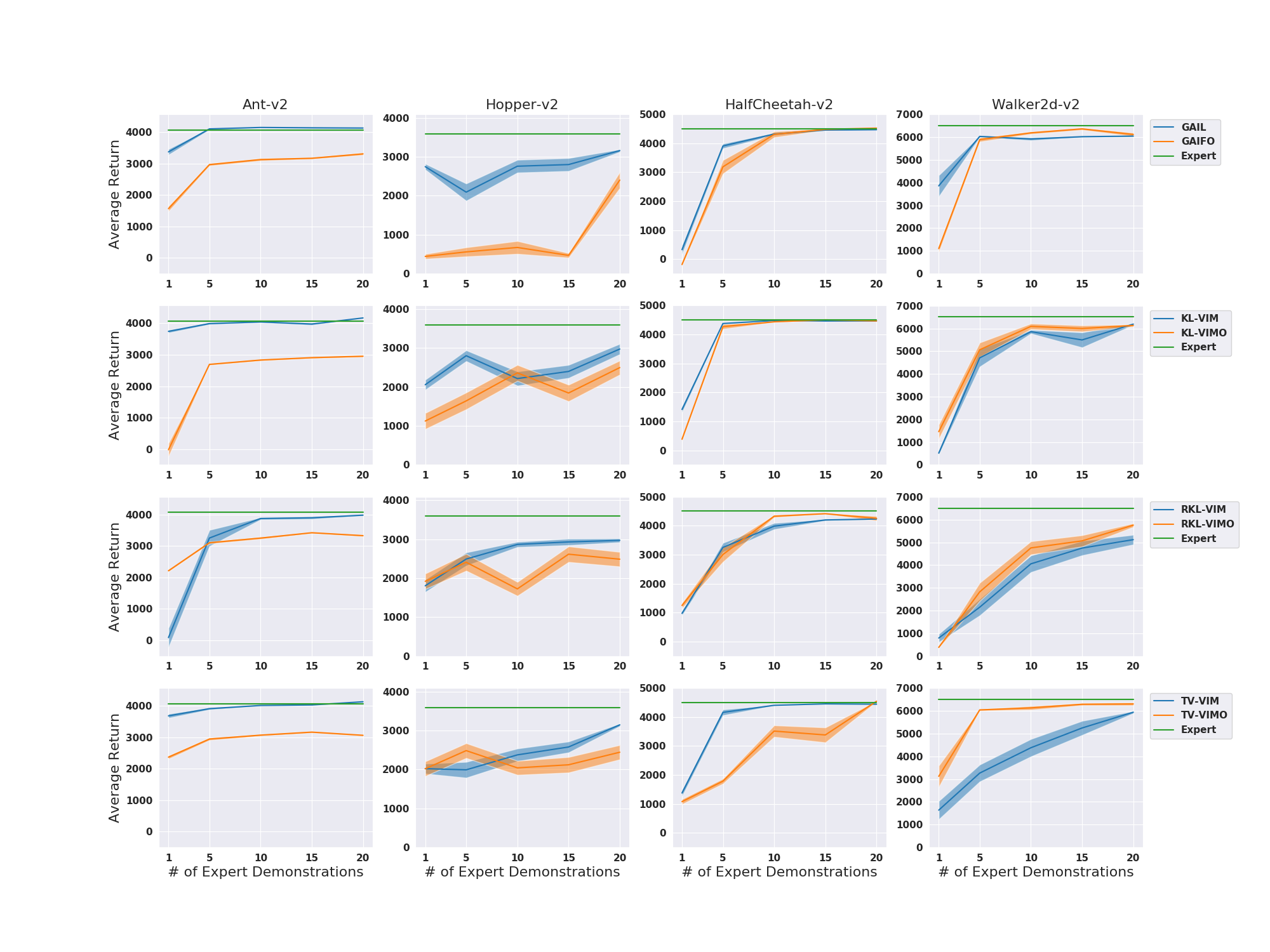}
    \caption{Evaluating $f$-VIM and $f$-VIMO across four MuJoCo environments with varying amounts of expert demonstration data.}
\end{figure*}

\newpage
\subsection{Sample Complexity Learning Curves}

\begin{figure*}[h!]
    \centering
    \includegraphics[width=\linewidth]{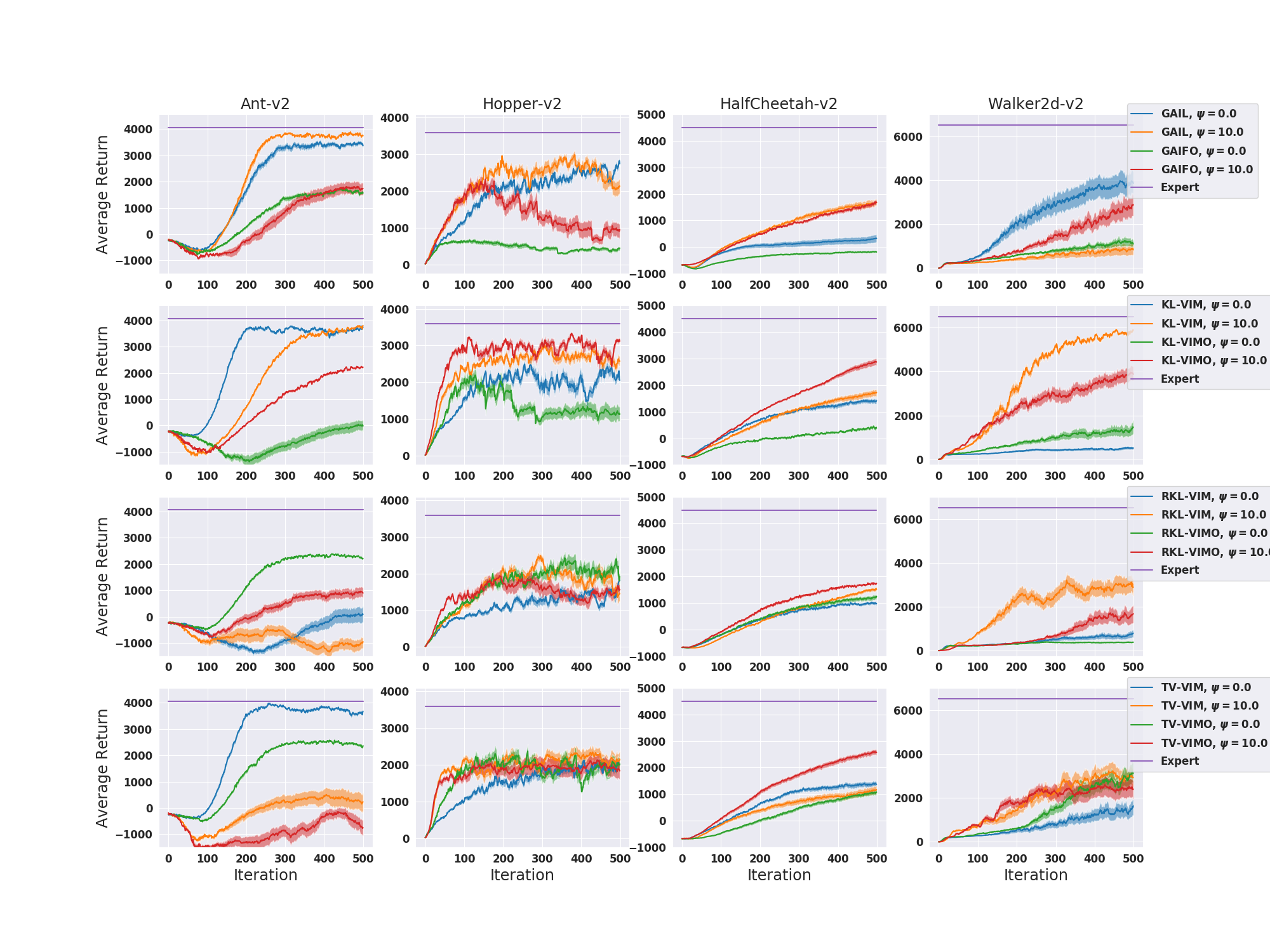}
    \caption{Learning curves for $f$-VIM and $f$-VIMO across four MuJoCo environments using only 1 expert demonstration.}
\end{figure*}

\newpage

\begin{figure*}[h!]
    \centering
    \includegraphics[width=\linewidth]{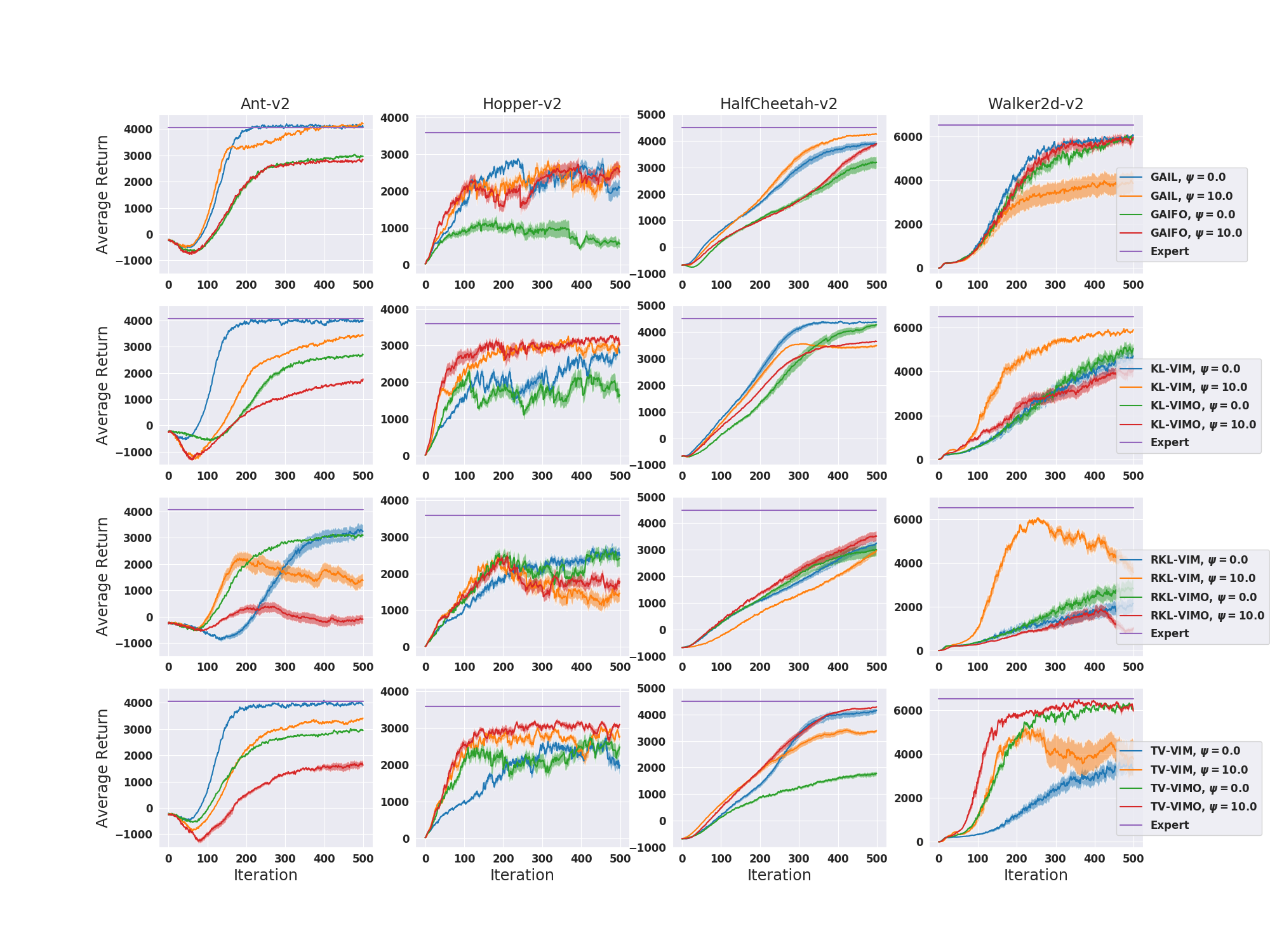}
    \caption{Learning curves for $f$-VIM and $f$-VIMO across four MuJoCo environments using only 5 expert demonstrations.}
\end{figure*}

\newpage

\begin{figure*}[h!]
    \centering
    \includegraphics[width=\linewidth]{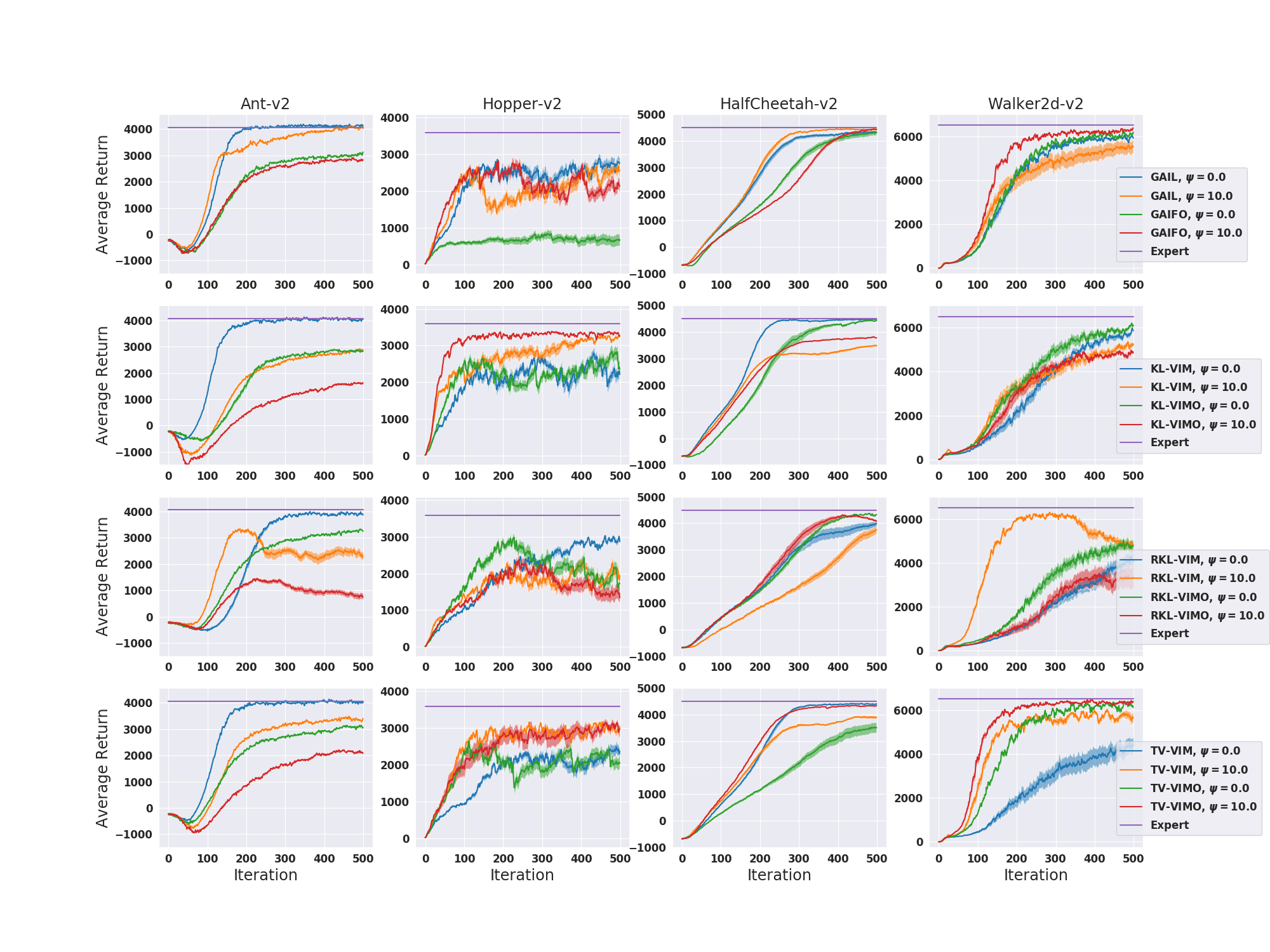}
    \caption{Learning curves for $f$-VIM and $f$-VIMO across four MuJoCo environments using only 10 expert demonstrations.}
\end{figure*}

\newpage

\begin{figure*}[h!]
    \centering
    \includegraphics[width=\linewidth]{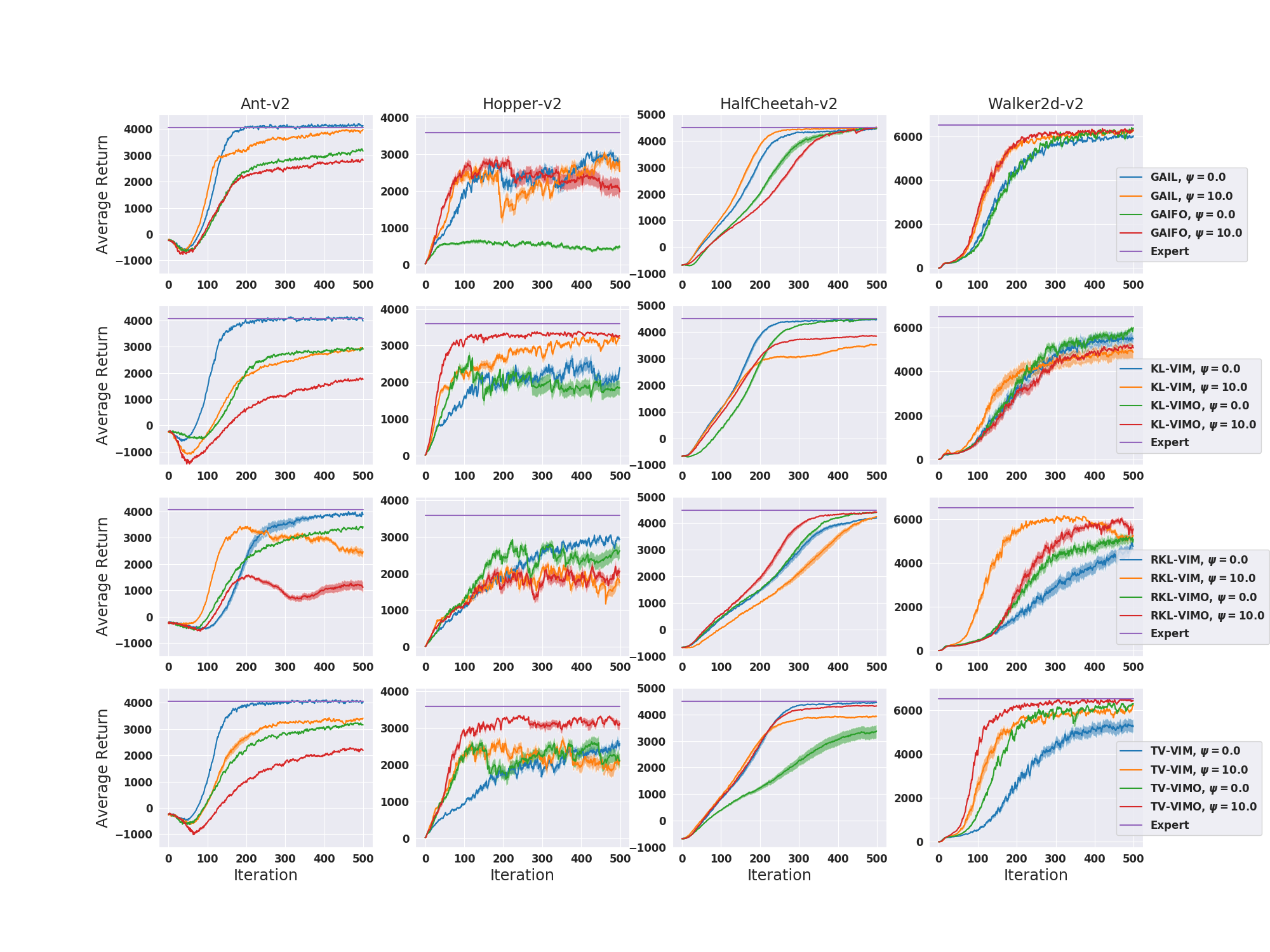}
    \caption{Learning curves for $f$-VIM and $f$-VIMO across four MuJoCo environments using only 15 expert demonstrations.}
\end{figure*}

\newpage

\begin{figure*}[h!]
    \centering
    \includegraphics[width=\linewidth]{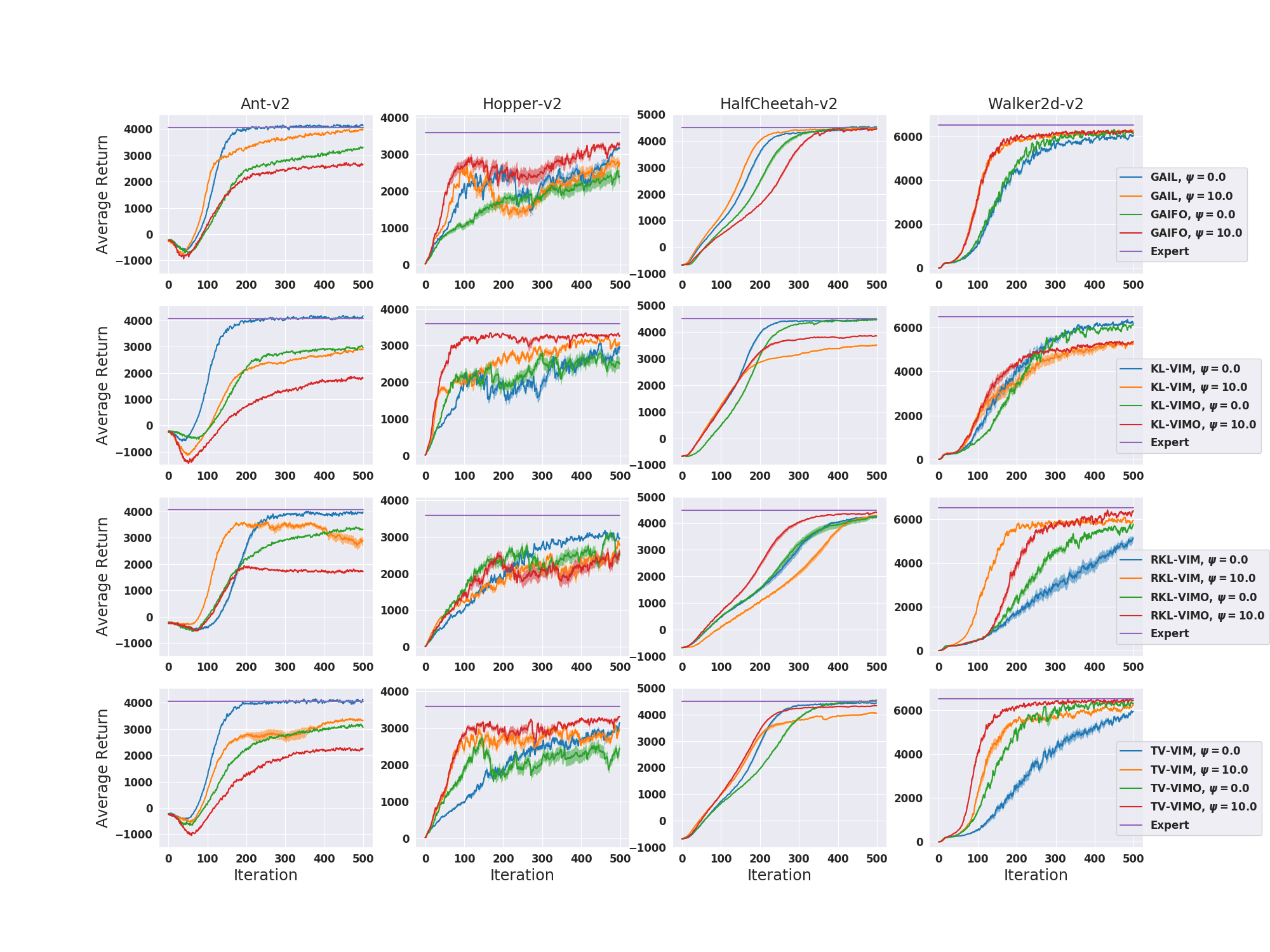}
    \caption{Learning curves for $f$-VIM and $f$-VIMO across four MuJoCo environments using only 20 expert demonstrations.}
\end{figure*}

\newpage

\section{$f$-Divergence Variational Bound Swap}
\label{sec:bound_swap}

Throughout this paper, we advocate for the use of the following variational lower bound to the $f$-divergence for both $f$-VIM and $f$-VIMO:
\begin{align}
    D_f(\rho^{\pie}||\rho^{\pi_\theta}) \geq \min\limits_\theta \max\limits_\omega \mathbb{E}_{(s,s') \sim \rho^{\pie}}[f^{*-1}(r(V_\omega(s,s')))] - \mathbb{E}_{(s,s') \sim \rho^{\pi_\theta}}[r(V_\omega(s, s'))]
    \label{eq:fvimo_supp}
\end{align}

In particular, we value the above form as it clearly exposes the choice of reward function for the imitation policy as a free parameter that, in practice, has strong implications for the stability and convergence of adversarial IL/ILO algorithms. Alternatively, one may consider appealing to the original lower bound of \citet{nguyen2010estimating}, used in $f$-GANs~\citep{nowozin2016f} unmodified, but swapping the positions of the two distributions:
\begin{align}
    D_f(\rho^{\pi_\theta}||\rho^{\pie}) \geq \min\limits_\theta \max\limits_\omega \mathbb{E}_{(s,s') \sim \rho^{\pi_\theta}}[g_f(V_\omega(s,s'))] - \mathbb{E}_{(s,s') \sim \rho^{\pie}}[f^*(g_f(V_\omega(s, s')))]
    \label{eq:fvim_obj_swap}
\end{align}

Consequently, the term in this lower bound pertaining to the imitation policy is now similar to that of the bound in Equation \ref{eq:fvimo_supp}; namely, an almost arbitrary activation function, $g_f$, applied to the output of the variational function (discriminator) $V_\omega$. The difference being that the codomain of $g_f$ must obey the domain of the convex conjugate, $f^*$, while the codomain of $r$ must respect the domain of the inverse convex conjugate, $f^{*-1}$. Moreover, while our reparameterization results in $r(s,a,s') = r(V_\omega(s,s'))$, the resulting per-timestep rewards under the swapped bound are given by $r(s,a,s') = -g_f(V_\omega(s,s'))$. In order to avoid repeating the dissipating signal issue observed with the Total Variation distance and keep consistent with sigmoid rewards, we select $g_f(u) = -\sigma(u)$ in our experiments.

We evaluate these two choices empirically below for the specific choice of the KL-divergence in the Ant and Hopper domains (the two most difficult domains of our evaluation). We find that the original unswapped bound in Equation \ref{eq:fvimo_supp} used throughout this paper outperforms the variants with the distributions swapper, for both the IL and ILO settings. Crucially, we find that the KL-VIM in the Ant domain no longer achieves expert performance while optimizing the swapped bound.

\begin{figure*}[h!]
    \centering
    \includegraphics[width=.75\linewidth]{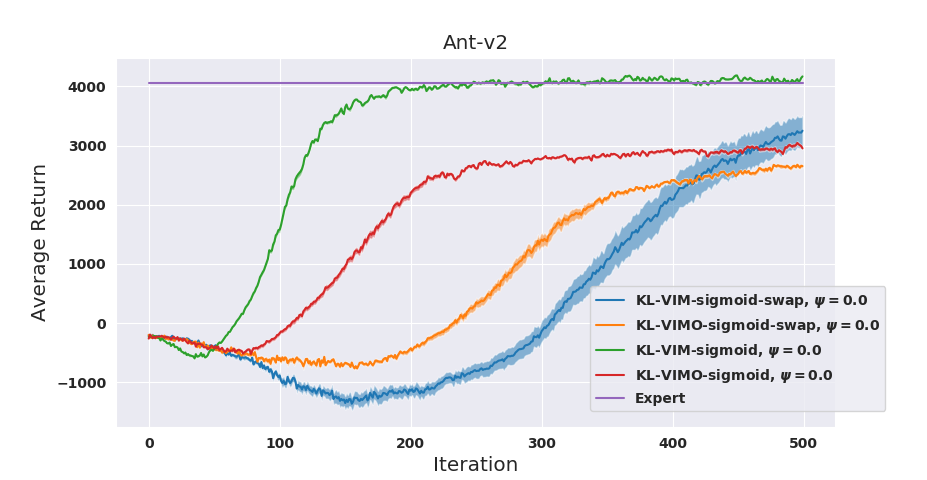}
    \caption{Learning curves for KL-VIM and KL-VIMO in Ant with 20 expert demonstrations using the regular and swapped variational lower bound.}
\end{figure*}

\begin{figure*}[h!]
    \centering
    \includegraphics[width=.75\linewidth]{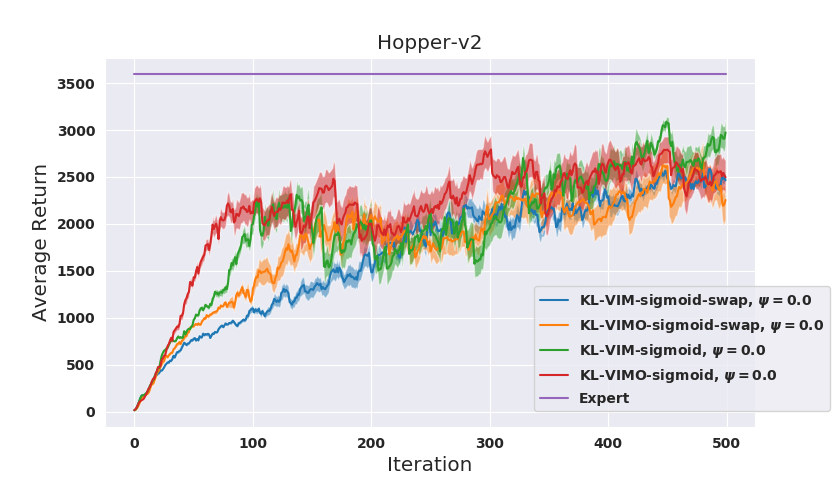}
    \caption{Learning curves for KL-VIM and KL-VIMO in Hopper with 20 expert demonstrations using the regular and swapped variational lower bound.}
\end{figure*}

\newpage
\bibliographystyle{icml2020}
\bibliography{references}






\end{document}